\documentclass[lettersize,journal]{IEEEtran}
\usepackage{amsmath,amsfonts}
\usepackage{algorithm}
\usepackage{array}
\usepackage[caption=false,font=normalsize,labelfont=sf,textfont=sf]{subfig}
\usepackage{textcomp}
\usepackage{stfloats}
\usepackage{url}
\usepackage{verbatim}
\usepackage{graphicx}
\usepackage{cite}
\usepackage{algpseudocode}
\usepackage{booktabs}
\usepackage{amsmath, amsfonts, amssymb}
\usepackage{multirow}
\usepackage[table,xcdraw]{xcolor}
\usepackage{pifont}
\usepackage{cleveref}
\Crefname{figure}{Fig.}{Figs.}
\Crefname{table}{TABLE}{TABLE}

\usepackage{soul}
\soulregister{\cite}7
\soulregister{\ref}7
\soulregister{\Cref}7
\soulregister{\includegraphics}7
\soulregister{\caption}7
\soulregister{\textbf}7
\soulregister{\textcolor}7
\soulregister{\footnote}7

\begin{document}

\title{CPG-PAD: Concept-Informed Prompts Guided Presentation Attack Detection}

\author{
    Haoyuan Zhang,
    Xiangyu Zhu, \IEEEmembership{Senior Member, IEEE},
    Li Gao,
    Ajian Liu,
    Siran Peng,
    Zhen Lei, \IEEEmembership{Fellow, IEEE}
    \thanks{Haoyuan Zhang is with the School of Artificial Intelligence, University of Chinese Academy of Sciences, Beijing 100049, China; the State Key Laboratory of Multimodal Artificial Intelligence Systems, Institute of Automation, Chinese Academy of Sciences, Beijing 100190, China  (e-mail: zhanghaoyuan2023@ia.ac.cn).}
    \thanks{Xiangyu Zhu, Ajian Liu and Siran Peng are with the State Key Laboratory of Multimodal Artificial Intelligence Systems, Institute of Automation, Chinese Academy of Sciences, Beijing 100190, China; the School of Artificial Intelligence, University of Chinese Academy of Sciences, Beijing 100049, China (e-mails: xiangyu.zhu@ia.ac.cn, ajian.liu@ia.ac.cn, pengsiran2023@ia.ac.cn).}
	\thanks{Li Gao is with the China Mobile Financial Technology Co., Ltd., Beijing 100032, China (e-mail: gaolids@chinamobile.com).}
	\thanks{Zhen Lei is with the State Key Laboratory of Multimodal Artificial Intelligence Systems, Institute of Automation, Chinese Academy of Sciences, Beijing 100190, China; the School of Artificial Intelligence, University of Chinese Academy of Sciences, Beijing 100049, China; the Centre for Artificial Intelligence and Robotics, Hong Kong Institute of Science and Innovation, Chinese Academy of Sciences, Hong Kong, China; the School of Computer Science and Engineering, the Faculty of Innovation Engineering, Macau University of Science and Technology, Macau, China (e-mail: zhen.lei@ia.ac.cn).}
    \thanks{Corresponding author: Zhen Lei.}
}

\markboth{Journal of \LaTeX\ Class Files,~Vol.~14, No.~8, August~2021}%
{Shell \MakeLowercase{\textit{et al.}}: A Sample Article Using IEEEtran.cls for IEEE Journals}

\maketitle

\begin{abstract}
Presentation Attack Detection (PAD) serves as a crucial safeguard for face recognition systems against presentation attacks such as printed photos, replayed videos, and 3D masks. Despite significant progress, existing PAD models still struggle to generalize across unseen domains due to variations in sensors, lighting, and attack materials. Recent Vision-Language Models (VLMs) have shown strong generalization ability, yet their applications in PAD remain limited because learned prompts, typically optimized under class-label supervision, fail to explicitly align with fine-grained attack-relevant visual semantics. As a result, the learned representations often overfit domain-specific artifacts instead of capturing transferable attack cues. To address this, we propose Concept-Informed Prompts Guided Presentation Attack Detection (CPG-PAD), a framework that introduces model-level concept guidance into the prompt learning process. Specifically, we design a Visual Concept-driven Enhancement (VCE) module that employs eXplainable AI (XAI) techniques to automatically discover PAD-relevant visual concepts and generate concept-associated heatmaps providing localized fine-grained guidance. Guided by these heatmaps, a Prompt-based Concept Injection (PCI) mechanism integrates these concepts into the prompt space through a Visual-Prompt Decoder (VPD) and a concept-mapping loss, enabling prompts to align with the model’s internal concept space. This design enables CPG-PAD to capture generalizable and domain-invariant attack cues while effectively suppressing dataset-specific biases. Extensive experiments across nine benchmark datasets demonstrate that CPG-PAD consistently achieves state-of-the-art cross-domain performance under multi-source, limited-source, and single-source settings.
\end{abstract}


\section{Introduction}

With the widespread adoption of face recognition in applications like identity verification and online payments, security concerns have become increasingly prominent. Presentation Attack Detection (PAD) is essential for detecting presentation attacks (PAs) using printed images \cite{print}, video replays \cite{replay}, and 3D masks \cite{mask}. While existing PAD methods, ranging from hand-crafted features \cite{kim2012face, yang2013face, zhang2011face} to deep-learning-based approaches \cite{flip, cfpl}, have shown promising results in controlled settings, they struggle to generalize across different domains due to significant domain shifts. Mitigating domain shifts between seen source samples and unseen target samples remains a considerable challenge for PAD models.

\begin{figure}[t]
    \centering
    \includegraphics[width=\columnwidth]{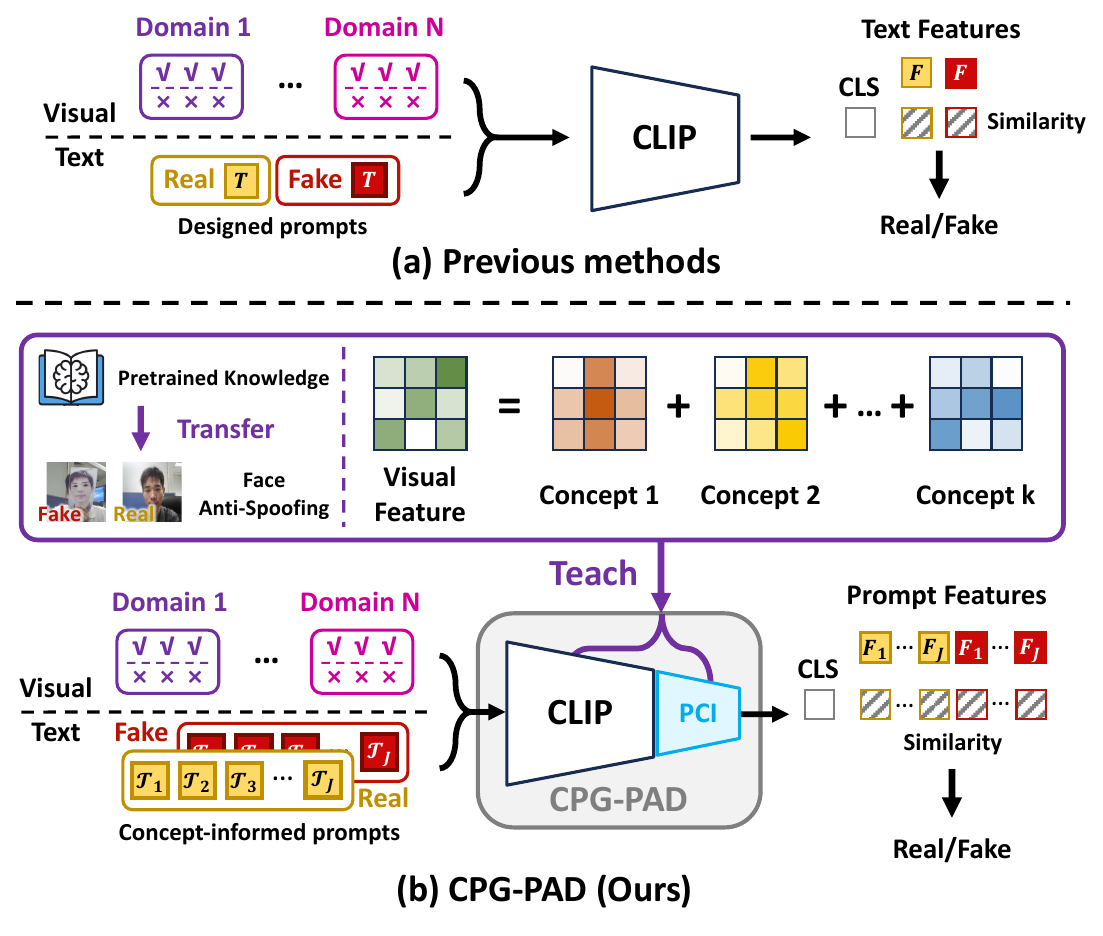}
    \caption{Comparison with Existing CLIP-like PAD Methods. (a) Previous methods aim to learn carefully designed prompts through class label supervision, which limits their performance. (b) In contrast, PAD-relevant visual concepts discovered from VLMs teach model to learn concept-informed prompts, which enhance PAD performance.}
    \label{fig:front}
\end{figure}

Traditional PAD methods \cite{iadg, gacfas, drcn} seek to bridge the distribution gap between source and target domains using adversarial adaptation, meta-learning, or feature alignment, etc. However, their performance is constrained by the uni-modal nature of these approaches. With the emergence of pretrained vision-language models (VLMs), the generalization of downstream tasks \cite{clip} improves through their rich multi-modal knowledge. Recent studies \cite{flip, cfpl} further show that CLIP-like frameworks significantly enhance PAD performance. These approaches outperform traditional PAD methods by providing additional context. However, human-level prompts struggle to convey the critical visual clues for PAD classification. For instance, clues like photo edges and moiré patterns can be described using natural language, while subtle geometric distortions and illumination-reflection discrepancies are difficult to express. As a result, constructing and learning prompts relying on human-level understanding limit the performance of existing CLIP-like PAD methods.

Surprisingly, we find that leveraging eXplainable AI (XAI) methods \cite{gradcam, ace, craft} enables us to transfer general knowledge embedded in pretrained VLMs into PAD process from a model-level perspective and feature space. This finding motivates us to construct and learn model-oriented prompts informed by these explanatory model-level signals. Specifically, with the aid of concept-based XAI techniques, we successfully discover visual concepts within the model's feature space and enhance domain data using corresponding feature heatmaps. These concept-associated heatmaps enable PAD models to learn multiple model-oriented prompts that are aligned with the discovered visual concepts, which guide PAD model to capture critical clues for classification beyond human-level understanding. A comparison with existing CLIP-like PAD methods is illustrated in \Cref{fig:front}.

Based on the above, we propose Concept-informed Prompts Guided Presentation Attack Detection (CPG-PAD), which exploits and incorporates visual concepts from pretrained VLMs into PAD process. Specifically, we introduce Visual Concept-driven Enhancement (VCE), a mechanism that leverages general pretrained VLMs to discover PAD-relevant visual concepts, while enhancing the domain data by generating fine-grained feature heatmaps corresponding to each discovered concept. The discovered concepts are closely linked to the objectives of attack detection, from a model-level perspective and within its feature space. With the aid of these concept-associated heatmaps, we introduce Prompt-based Concept Injection (PCI) to learn multiple concept-informed prompts by aligning textual prompts with the discovered visual concepts. In particular, learnable prompts interact with visual features through the Visual-Prompt Decoder (VPD) and are supervised by concept-associated heatmaps to inject concept information into the prompts. In this way, CPG-PAD accurately captures generalizable visual features, suppressing interference from domain-specific information when encountering unseen target domains, thereby enhancing the generalizability of PAD task.

Our contributions can be summarized as follows:
\begin{itemize}
    \item We propose Concept-informed Prompts Guided PAD (CPG-PAD), a novel framework which enhances domain generalization performance by incorporating visual concepts discovered by employing XAI techniques on pretrained VLMs.
    \item CPG-PAD introduces Visual Concept-driven Enhancement (VCE), and Prompt-based Concept Injection (PCI). VCE automatically discovers PAD-relevant visual concepts, and enhances domain data by generating corresponding feature heatmaps. With the concept-associated heatmaps, PCI learns multiple concept-informed prompts through Visual-Prompt Decoder (VPD).
    \item Extensive experiments and analysis demonstrate the superiority of CPG-PAD over state-of-the-art competitors on widely-used benchmark datasets.
\end{itemize}

\section{Related Works}

\subsection{Presentation Attack Detection}

The earliest attempts at Presentation Attack Detection (PAD) focused on handcrafted feature engineering. Techniques such as Local Binary Patterns (LBP) \cite{lbptop, lbp2}, Histogram of Oriented Gradients (HOG) \cite{hog}, and Scale-Invariant Feature Transform (SIFT) \cite{sift} were employed to capture texture irregularities caused by presentation attacks. Although these methods provided a foundation for early PAD, they were easily affected by changes in environmental conditions, including lighting variations, scaling factors, and head pose, which hindered robustness in practice.

Deep learning advanced PAD research by enabling robust and discriminative feature extraction. Convolution Neural Networks (CNNs) quickly became the mainstream choice \cite{cnn1, cnn2, cnn3}. Yet, the inherently local nature of convolutions limited their capacity to jointly capture fine-grained details and long-range dependencies, leading to vulnerabilities in challenging attack scenarios \cite{tf1}. To overcome these shortcomings, transformer-based models \cite{tf2, tf3} have been explored. By leveraging self-attention mechanisms, they model global relationships across image patches and extend beyond CNNs’ receptive field constraints. While these architectures have achieved strong intra-domain evaluation, they often fail to generalize well in unseen environments.

To address the domain shift problem, researchers have explored Domain Adaptation (DA) and Domain Generalization (DG) strategies. DA \cite{druda, sda, udgfas} aims to bridge distribution gaps between source and target domains using unlabeled target data. Existing methods employ adversarial adaptation \cite{druda}, meta-learning \cite{sda}, or extract generalizable features from large-scale unlabeled data \cite{udgfas}. However, obtaining target domain data during training remains a significant challenge in real-world scenarios. In contrast, DG \cite{drdg, rfmeta, iadg, ttdg, gacfas} tackles this issue by enhancing model robustness across multiple source domains without relying on target data. Traditional DG approaches learn domain-invariant features through feature alignment \cite{iadg} or hierarchical relationships \cite{hpdr} to improve generalizability in PAD. Despite these improvements, single-modality DG-PAD still faces difficulties when domain discrepancies are large or training data limited.

Unlike DA and DG, which focus on improving feature generalization, auxiliary-supervised approaches provide additional supervisory signals to guide feature learning. Some works augment the classification task with physical or physiological auxiliary signals such as depth maps or remote photoplethysmography (rPPG). For example, Liu et al. \cite{cnn1} jointly estimate pixel-wise facial depth and rPPG rhythm along with the binary real/fake decision, showing that these auxiliary losses help guide the network to focus on fundamental attack-discriminative cues rather than overfit to texture patterns. Subsequent works \cite{cnn3} refine this supervision by introducing spatial gradient features and contrastive depth losses to sharpen the depth supervision signal. Other methods \cite{aux1, aux2, aux3} extend depth supervision across multiple frames or design adaptive fusion with rPPG, e.g. exploiting temporal and depth information using multiframe depth estimation. Related efforts also introduce explanation-based or attention-based guidance as auxiliary supervision, using saliency maps to regularize model attention and improve interpretability \cite{pan2022}.

Recent progress in large-scale pretrained vision and vision-language models, particularly CLIP \cite{clip}, has opened new avenues for PAD. A first line of research \cite{foundation_model_pad, foundpad} investigates how foundation models can be adapted to PAD through simple transfer strategies, such as feature extraction, lightweight adaptation, or zero-shot inference. These studies suggest that pretrained models provide strong general visual priors, but directly transferring such knowledge to PAD remains challenging without task-specific guidance. Building on this, CLIP-based PAD methods further exploit cross-modal supervision by aligning facial images with textual prompts. FLIP \cite{flip} fine-tunes pretrained vision-language models by associating facial images with textual prompts such as ``a photo of a real face'' or ``a photo of a fake face'', achieving substantial improvements in DG-PAD. Building on this, subsequent works \cite{cfpl, scptl, ccpe, tffas, dgpdl, icpe} propose frameworks that more effectively leverage multi-modal knowledge by disentangling features or constructing fine-grained prompts. Although existing CLIP-like methods achieve better performance than traditional PAD methods, their generalizability is still constrained by domain-specific biases.

\subsection{Visual Concept Discovery in XAI}

Explainable AI (XAI) aims to improve the interpretability and transparency of deep learning models by providing explanations for their decisions. In computer vision, gradient-based methods \cite{gradcam, ablationcam, gradcampp, lrp} leverage activation and gradient information to attribute decisions to specific image regions. Perturbation-based methods \cite{rise} perturb the input image and observe the resulting changes in the model’s output to infer decision attribution. However, these methods focus on attributing individual decisions locally and lack the ability to explain the model from a global perspective. In contrast, concept-based approaches aim to generate multiple global explanations by analyzing groups of data that share common characteristics. Kim et al. \cite{tcav} first introduced a method that goes beyond attribution-based approaches by measuring the influence of pre-selected concepts on a model’s outputs. ACE \cite{ace} and CRAFT \cite{craft} further advance this idea by automatically discovering concepts through clustering and factorization of intermediate neural activations. Unfortunately, existing concept-based XAI methods are primarily designed for general classification tasks and are not well suited to specific downstream applications. Building on these ideas, we seek to adapt concept-based XAI techniques for PAD, using visual concepts as a novel source of supervision.

\subsection{Motivation}

In the history of PAD research, auxiliary supervision from signals such as rPPG, depth maps, and NIR images has been shown to improve domain generalization by providing complementary cues beyond RGB data. At the same time, explainability techniques enable us to uncover machine-level concepts and heatmaps from powerful pretrained models, offering a new form of supervision that does not rely on additional sensors. This perspective differs from prior attention-guided PAD methods \cite{pan2022} that provide sample-level saliency supervision, as well as recent CLIP-based methods such as TF-FAS \cite{tffas} and CCPE \cite{ccpe} that mainly transfer pretrained knowledge through richer semantic guidance or prompt engineering. Motivated by this, we propose a Visual Concept-driven Enhancement (VCE) process to distill PAD-relevant concepts and heatmaps from the CLIP visual encoder. These concepts and heatmaps are incorporated into a novel learning framework as auxiliary supervision signals, which are injected into learnable prompts to strengthen the generalization ability of PAD models.

\begin{figure*}[t]
    \centering
    \includegraphics[width=\textwidth]{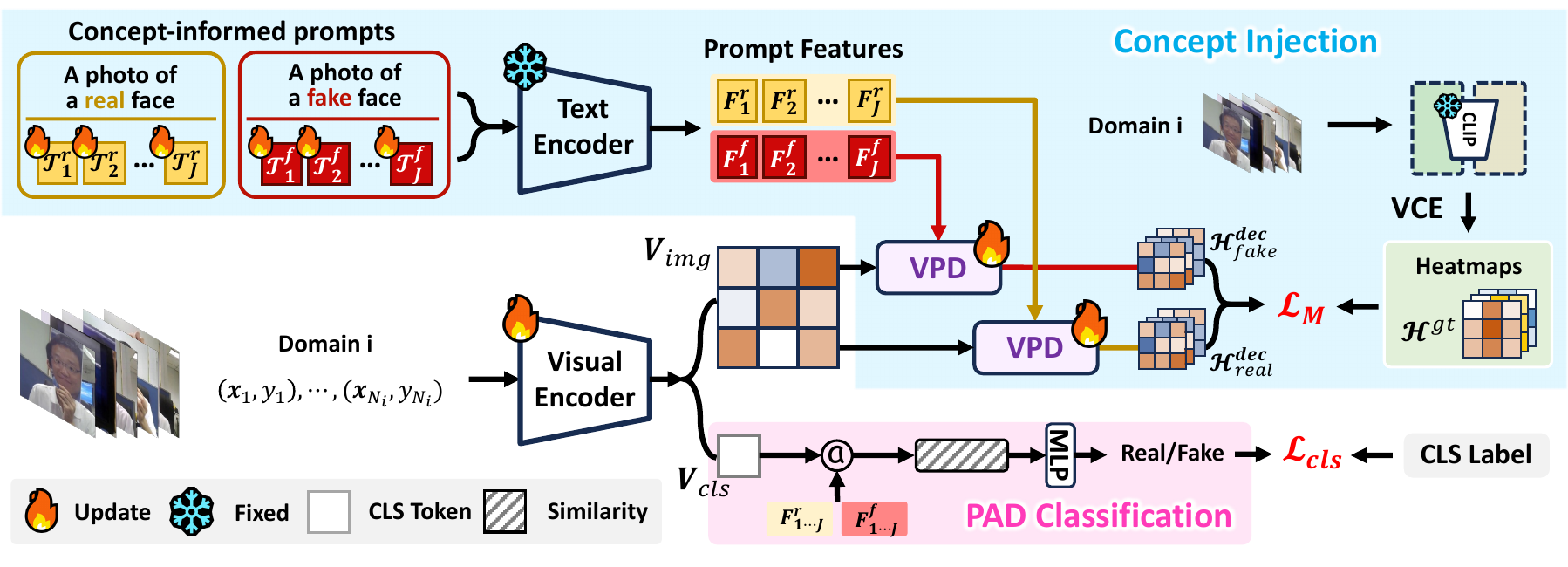}
    \caption{Detailed information of Concept-informed Prompts Guided PAD (CPG-PAD). We first generate concept-associated heatmaps $\mathcal{H}^{gt}$ through Visual concept-driven Enhancement (VCE) (details in \Cref{fig:vce}). Next, multiple concept-informed prompts are learned via Prompt-based Concept Injection (PCI). Specifically, domain visual data is encoded into a CLS token $V_{cls}$ and image tokens $V_{img}$ through a learnable visual encoder. Learnable prompt embeddings $\{\mathcal{T}^{r/f}_j\}_{j=1}^J$ and fixed class-wise prompts ``A photo of a \{CLS\} face" is encoded into Prompt Features $\{F^{r/f}_j\}_{j=1}^J$ through a fixed text encoder. Multiple feature heatmaps $\mathcal{H}^{dec}$ are decoded through Visual-Prompt Decoder (VPD) by interaction between $\{F^{r/f}_j\}_{j=1}^J$ and $V_{img}$. The training process is supervised by the Cross Entropy loss $\mathcal{L}_{cls}$ and a concept mapping loss $\mathcal{L}_M$, which injects visual concept into learnable prompts by aligning $\mathcal{H}^{dec}$ with $\mathcal{H}^{gt}$.}
    \label{fig:method}
    \vspace{-0.2cm}
\end{figure*}

\section{Method}

The proposed Concept-informed Prompts Guided PAD (CPG-PAD) can be decomposed into two steps. We first enhance each domain through Visual Concept-driven Enhancement (VCE). Then we learn multiple concept-informed prompts via Prompt-based Concept Injection (PCI) process. The total framework of CPG-PAD is shown in \Cref{fig:method}.

\subsection{Preliminaries}

We adopt a general domain generalization PAD training setting. $\mathcal{D}$=$\{D_i, i = 1, 2, \cdots, I\}$ denotes the domain data where the $i\mbox{-}th$ domain $\mathcal{D}_i = \{(\boldsymbol{x}_j, y_j), j=1, 2, \cdots, N_i\}$ contains $N_i$ image and label pairs. $\mathcal{C}(\cdot)$ denotes CLIP model with pretrained parameters. CLIP model $\mathcal{C}(\cdot)$ contains two parts, the visual encoder $VisEnc(\cdot)$ extracts CLS token $V_{cls} \in \mathbb{R}^{d_v}$ and image tokens $V_{img} \in \mathbb{R}^{H \times W \times d_v}$ and the text encoder $TextEnc(\cdot)$ generates text feature $F \in \mathbb{R}^{d_t}$ for each category:
\begin{align}
    V_{cls}, V_{img} = VisEnc(\mathcal{X}); \ \ \ F = TextEnc(\mathcal{T}) \nonumber
\end{align}
where $\mathcal{X}$ and $\mathcal{T}$ indicate input image and prompts respectively. Finally, visual tokens $V_{cls}, V_{img}$ and text feature $F$ are projected to the same dimension $C$ to calculate similarity scores. Notably, $d_t=C=512$, $d_v=768$ in ViT-B/16.

\subsection{Visual Concept-driven Enhancement}

\begin{figure}[t]
    \centering
    \includegraphics[width=\columnwidth]{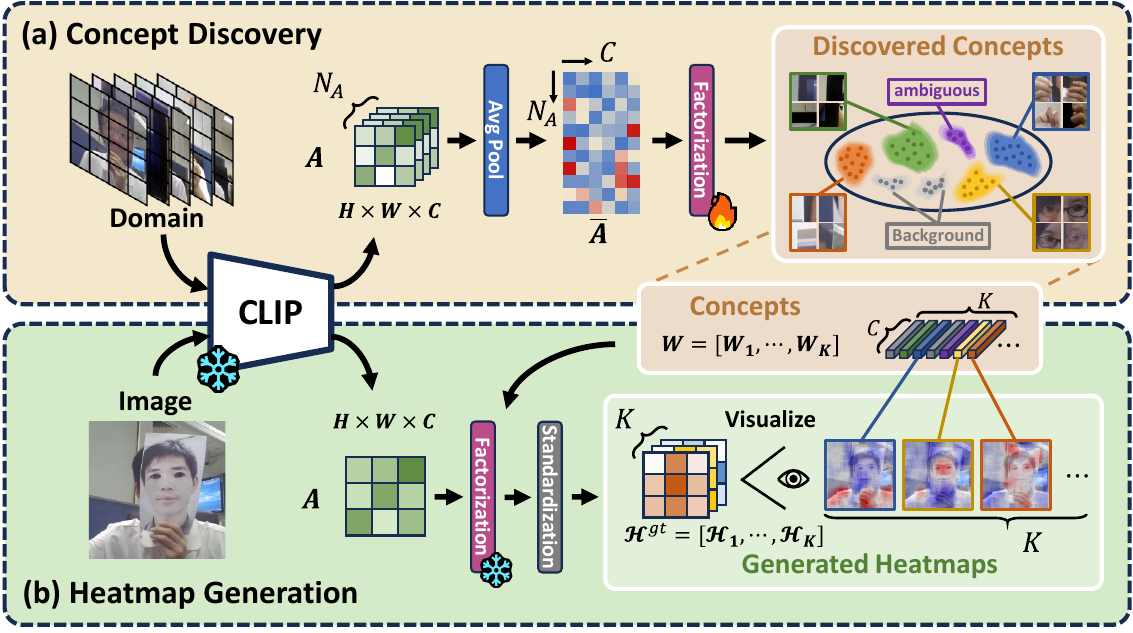}
    \caption{Detailed information of Visual Concept-driven Enhancement (VCE). VCE consists of two stages, (a) concept discovery and (b) concept-associated heatmap generation.}
    \label{fig:vce}
\end{figure}

In this section, we introduce Visual Concept-driven Enhancement (VCE), which operates in two stages: concept discovery and concept-associated heatmap generation. Notably, VCE utilizes a pretrained CLIP model without requiring any parameter updates. The detailed process is illustrated in \Cref{fig:vce}. For clear illustration, we use the symbol $N$ to represent the number of domain inputs for all domains. 

\subsubsection{\textbf{Concept Discovery}}

The process of concept discovery is illustrated in \Cref{fig:vce}(a) with \textcolor[RGB]{191,144,0}{\textbf{dark yellow box}}. We generally consider a single domain $\mathcal{D}_i$, the data can be simply represented as input images $(\boldsymbol{x}_1, \cdots, \boldsymbol{x}_{N}) \in \mathcal{X}^{N}$ where $\boldsymbol{x_j} \in \mathbb{R}^{D}$ and their associated labels $(y_1, \cdots, y_{N}) \in \mathcal{Y}^{N}$. Firstly, we select subsets of images $\mathcal{X}^{fake/real}_{sub} \in \mathbb{R}^{N_s \times D}$ from the original dataset $\mathcal{X}^{N}$ for both real and fake class to discover concepts. Here we randomly select $r$ frames from each video to form the subset where $r$ is a hyperparameter (e.g. select a single frame from each live video to form $\mathcal{X}^{real}_{sub}$). We define $\pi(\cdot)$ as a straightforward filter function to create candidate concepts. Specifically, $\pi(\cdot)$ crops the image into $64 \times 64$ patches at regular intervals in both vertical and horizontal directions and resize each patch to dimension $D$. Feed $\mathcal{X}_{sub}$ into $\pi(\cdot)$ to obtain an auxiliary dataset $\mathcal{X}^{fake/real}_{patch} \in \mathbb{R}^{N_a \times D}$ which contains all candidate concepts. The following part will focus on the auxiliary dataset of a single class $\mathcal{X}_{patch} \in \mathbb{R}^{N_a \times D}$, as the same operations apply to both the fake and real classes.

To automatically discover visual concepts from domain auxiliary dataset $\mathcal{X}_{patch}$, we feed it to the pretrained visual encoder to obtain activation $\mathbf{A} = VisEnc (\mathcal{X}_{patch}) \in \mathbb{R}^{N_a \times H \times W \times C}$ where $HW$ indicates the shape of activation map and $C$ indicates the number of channels. We apply Semi-NMF (Semi Non-negative Matrix Factorization) \cite{seminmf} to factorize activation maps, as it can naturally handle mixed-sign activations in CLIP features while encouraging interpretable, parts-based representations, which are difficult to achieve with other factorization methods. For instance, standard NMF is not applicable due to its non-negativity constraint on inputs, while PCA tends to produce less semantically interpretable components \cite{lens}. Semi-NMF decomposes the average pooled activations $\mathbf{\bar{A}} = AvgPool(\mathbf{A}) \in \mathbb{R}^{N_a \times C}$ into a product of concept coefficients $\mathbf{U} \in \mathbb{R}^{N_a \times K}$ and concept basis $\mathbf{W} \in \mathbb{R}^{C \times K}$ by solving:
\begin{align}
	(\mathbf{U}, \mathbf{W}) = \mathop{\arg\min}\limits_{\mathbf{U} \geq 0, \mathbf{W}} \Vert \mathbf{\bar{A}} - \mathbf{U} \mathbf{W}^\intercal \Vert^2_F, \label{eq:seminmf}
\end{align}
where $K$ indicates the number of concepts one wishes to discover and $ \Vert \cdot \Vert^2_F $ denotes the Frobenius norm.

According to Ding \textit{et al.} \cite{seminmf}, the objective can be solved by iteratively updating $\mathbf{U}$ and $\mathbf{W}$ as follows:
\begin{align}
	& \mathbf{W} = \boldsymbol{\bar{A}}^\intercal \mathbf{U} (\mathbf{U}^\intercal \mathbf{U})^{-1}, \\
	& \mathbf{U} = \mathbf{U} \cdot \sqrt{\frac{ (\boldsymbol{\bar{A}} \mathbf{W})^{+} + \mathbf{U} (\mathbf{W}^\intercal \mathbf{W})^{-} }{ (\boldsymbol{\bar{A}} \mathbf{W})^{-} + \mathbf{U} (\mathbf{W}^\intercal \mathbf{W})^{+} } },
\end{align}
where the separated parts of matrix $\mathbf{M}$ are:
\begin{align*}
	\mathbf{M}^{+} = (\vert \mathbf{M} \vert + \mathbf{M}) / 2, \ \ \ \mathbf{M}^{-} = (\vert \mathbf{M} \vert - \mathbf{M}) / 2
\end{align*}

$\mathbf{W}$ is the discovered concepts where each column $\mathbf{W}_k \in \mathbb{R}^{C}$ corresponds to a single concept basis. These concepts will be utilized in the subsequent heatmap generation process.

\subsubsection{\textbf{Concept-Associated Heatmap Generation}}

The process of concept-associated heatmap generation is illustrated in Figure~\ref{fig:vce}(b) with \textcolor[RGB]{79,126,50}{\textbf{green box}}. After concept discovery, we get $K$ concept basis $\mathbf{W} \in \mathbb{R}^{C \times K}$ from the selected subset $\mathcal{X}_{sub}$ with $N_s$ images, which serves as an approximation of the concepts present in the original domain containing $N$ images. Given an image $\boldsymbol{x}$ (from the original domain), we factorize the activations $\mathbf{A} = VisEnc(\boldsymbol{x}) \in \mathbb{R}^{H \times W \times C}$ in each position into several concept coefficients $\mathbf{U} \in \mathbb{R}^{H \times W \times K}$ corresponding to $\mathbf{W}$ by solving similar objective in Equation~\ref{eq:seminmf}:
\begin{align}
	(\mathbf{U}(s, t), \mathbf{W}) = \mathop{\arg\min}\limits_{\mathbf{U}(s, t) \geq 0, \mathbf{W}} \Vert \mathbf{A}(s, t) - \mathbf{U}(s, t) \mathbf{W}^\intercal \Vert^2_F, \label{eq:seminmf2}
\end{align}
where $\mathbf{W}$ is fixed based on the values obtained during the concept discovery process, and $\mathbf{A}(s, t) \in \mathbb{R}^{C}$ represents the activation vector at position $(s, t)$ within the spatial dimensions $(H, W)$. The result $\mathbf{U}(s, t)  \in \mathbb{R}^{K}$ indicates the concept coefficients in position $(s, t)$.

Specifically, we can regard $\mathbf{U}(s, t, k)$ as the influence of concept $k$ at $(s, t)$ position of the activation map and the $\mathbf{U}(\cdot, \cdot, k)$ can be seen as an attention map of concept $k$. In this way, we factorize a single input image into several concepts and their corresponding attention maps. With the help of $\mathbf{U}$, we can get the fine-grained feature heatmaps by standardizing them to have zero mean and unit variance:
\begin{align}
    \mathcal{H}^{gt}(k) = \frac{\mathbf{U}(\cdot, \cdot, k)- mean(\mathbf{U}(\cdot, \cdot, k))}{std(\mathbf{U}(\cdot, \cdot, k))},
\end{align}
where $mean$ and $std$ are operators to calculate the mean and variance of activation map $\mathbf{U}(\cdot, \cdot, k)$.

\subsection{Prompt-based Concept Injection}

In this section, we propose Prompt-based Concept Injection (PCI), which learns concept-informed prompts by integrating visual concept information into textual prompts. PCI constructs multiple learnable prompts $\mathcal{S}$ for both fake and real classes. For concept injection, $\mathcal{S}$ is decoded into heatmaps $\mathcal{H}^{dec}$ via a proposed Visual-Prompt Decoder (VPD) and supervised by concept-associated heatmaps $\mathcal{H}^{gt}$ generated by the former VCE to align with discovered visual concepts. For PAD classification, the well-learned concept-informed prompts $\mathcal{S}$ guide CPG-PAD to identify generalizable features and project to the final classification logits $l$.

\subsubsection{\textbf{Learnable Prompts}}

The learnable prompts $\mathcal{S}$ consist of two components: fixed class-wise text descriptions $\mathcal{P}^{text}$ and multiple learnable prompt embeddings $\{ \mathcal{T}^{prompt}_j \}^{J}_{j=1}$, where $J$ is a hyperparameter. Following the setup in \cite{flip}, we construct the fixed class-wise text description $\mathcal{P}^{text}$ as \{A photo of a fake face\} and \{A photo of a real face\}. These texts are tokenized and embedded using CLIP’s tokenizer and embedding layer, resulting in fixed class-wise embeddings $\mathcal{T}^{text} \in \mathbb{R}^{n_t \times d_t}$. The learnable prompt embeddings $\{ \mathcal{T}^{prompt}_j \}^{J}_{j=1}$ where $\mathcal{T}^{prompt}_j \in \mathbb{R}^{n_p \times d_t}$ are concatenated with the fixed class-wise embeddings $\mathcal{T}^{text}$ and the necessary $[SOS]$ and $[EOS]$ embeddings to form complete learnable prompts $\mathcal{S}$. The process can be formulated as follows:
\begin{align}
    & [SOS], \mathcal{T}^{text}, [EOS] = \mathcal{C}_{Embed}( \mathcal{C}_{tokenize}( \mathcal{P}^{text} ) ), \nonumber \\
    & \ \ \ \ \ \ \ where \ \   \mathcal{T}^{text} \in \mathbb{R}^{n_t \times d_t},\ \ [EOS], [SOS] \in \mathbb{R}^{d_t},\nonumber \\ 
    & \mathcal{S}_j = concat([SOS], \mathcal{T}^{prompt}_j, \mathcal{T}^{text}, [EOS]) \in \mathbb{R}^{77 \times d_t}, \nonumber \\
    & \mathcal{S} = \{ \mathcal{S}_j \}^{J}_{j = 1} \in \mathbb{R}^{J \times 77\times d_t}, \nonumber
\end{align}
where $\mathcal{C}$ indicates CLIP model, $d_t$ indicates embedding dimension of the text encoder in CLIP. The parameters $n_t$ and $n_p$ correspond to the embedding lengths of the fixed class-wise embeddings and the learnable prompt embeddings, respectively, ensuring that their sum satisfies $n_t + n_p = 75$. The complete learnable prompts $\mathcal{S}$ are subsequently aligned with visual concept information through concept injection. Notably, the same learnable prompts $\mathcal{S}^{r/f} =\mathcal{S}$ are constructed for both fake and real classes.

\begin{figure}[t]
    \centering
    \includegraphics[width=0.7\columnwidth]{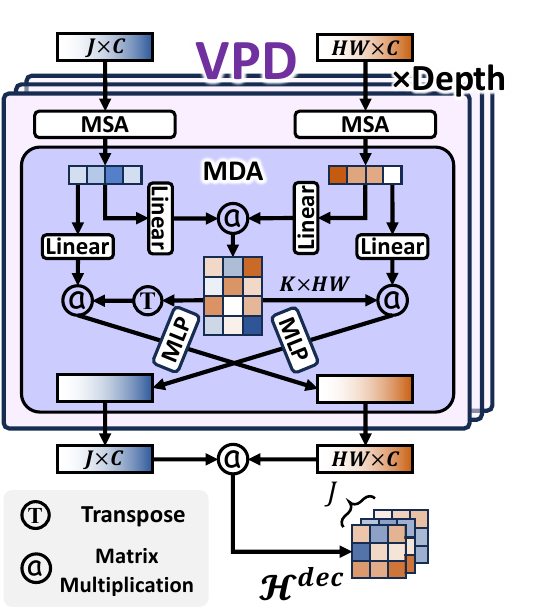}
    \caption{Detailed structure of Visual-Prompt Decoder (VPD) module. MSA denotes multi-head self attention and MDA denotes Multi-head dual attention.}
    \label{fig:vpd}
    \vspace{-0.4cm}
\end{figure}

\subsubsection{\textbf{Learning Prompts via Concept Injection}}

The prompt learning process via concept injection is illustrated in Figure~\ref{fig:method} with \textcolor[RGB]{0,191,255}{\textbf{blue box}}. With the constructed learnable prompts $\mathcal{S}^{r/f}$, we propose a Visual-Prompt Decoder (VPD) to better learn multiple concept-informed prompts for both fake and real classes. VPD can effectively decode multiple feature heatmaps with the interaction between multiple prompts $\mathcal{S}^{r/f}$ and image tokens $V_{img}$, which is later aligned with discovered concepts. For clear description, we use $\mathcal{S}$ to represent $\mathcal{S}^f$ or $\mathcal{S}^r$.

Feed the prompts $\mathcal{S}$ into the text encoder to obtain the prompt features $F = TextEnc(\mathcal{S}) \in \mathbb{R}^{J \times C}$. Similarly, process the input image through the image encoder to derive a CLS token and image tokens $V_{cls}, V_{img} = VisEnc(X)$ where $V_{cls} \in \mathbb{R}^{C}$ and $V_{img} \in \mathbb{R}^{H \times W \times C}$. VPD takes the prompt features $F$ and image tokens $V_{img}$ as inputs and decodes into heatmaps $\mathcal{H}^{dec} = VPD(F, V_{img})$ through their interactions. The decoded heatmaps are later supervised by concept-associated heatmaps $\mathcal{H}^{gt}$.

Specifically, VPD is a module with several layers of multi-head self-attention mechanism $MSA(\cdot)$ (identical to the traditional Transformer \cite{transformer}) and a newly proposed multi-head dual-attention module $MDA(\cdot)$. Multi-head dual-attention calculates a single attention map for the two inputs and updates both through matrix multiplication to enable effective interaction, illustrated in Figure~\ref{fig:vpd}. For each layer in VPD, it first applies multi-head self-attention to both the image tokens $ V_{img} $ and the prompt features $ F $, producing $\hat{V}_{img}$ and $\hat{F}$, respectively. Then, the processed image tokens $\hat{V}_{img}$ and prompt features $\hat{F}$ are projected into query $Q_*$ and key $K_*$ representations using separate linear layers with layer normalization, resulting in $ Q_V, K_V $ for image tokens and $ Q_F, K_F $ for prompt features where $ Q_V, K_V \in \mathbb{R}^{H \times W \times C} $ and $ Q_F, K_F \in \mathbb{R}^{J \times C} $. Next, the attention matrix $Attn$ is computed using a scaled dot-product between $ Q_V $ and $ Q_F $, followed by a softmax operation forming the shape $\mathbb{R}^{H \times W \times J} $. The final representations are obtained by applying two separate Multi-Layer Perceptrons (MLPs) to the attended features. The image tokens are updated using the attended prompt keys $K_F$, while the prompt features are refined through the attended image keys $K_V$. Notably, the MLPs consist of two linear layers with GELU activation.

After passing through the Depth layer of the VPD, the decoded heatmap $\mathcal{H}^{dec} \in \mathbb{R}^{H \times W \times J}$ is obtained via $ \mathrm{Conv2d} \left(F @ {V_{img}}^\intercal\right) $. With the generated heatmaps $\mathcal{H}^{gt}$, we inject visual concept information into learnable prompts through concept mapping loss $\mathcal{L}_M$ which supervises the class-specific heatmaps conditioned on the ground-truth label by minimizing the optimal Mean Square Error (MSE) through Hungarian Matching \cite{hungarian}:
\begin{gather}
    \mathcal{H}^{dec} = (1-\delta)\mathcal{H}^{dec}_{fake} + \delta\mathcal{H}^{dec}_{real}, \text{where } \delta = \mathbb{I}(\text{label=real}), \nonumber \\
    \mathcal{L}_M = \frac{1}{J} \min_{\pi \in S_J} \sum_{j=1}^{J} MSE(\mathcal{H}^{dec} (j), \mathcal{H}^{gt}(\pi(j))).
\end{gather}
where $\mathbb{I}(\cdot)$ is the indicator function, $S_J$ is the set of all permutations of $\{1, \dots, J\}$, $\pi$ is the optimal permutation found via the Hungarian algorithm.

Thus, we learn the concept-informed prompts $\mathcal{S}^{r/f}$ through visual concept injection with the help of the VPD.

\subsubsection{\textbf{PAD Classification}}

With the learned concept-informed prompts, CPG-PAD can identify significant features in PAD classification process, illustrated in \Cref{fig:method} with pink dotted box. Specifically, prompt features $F^{r/f}$ are used to calculate the similarity score $s = concat(F^r @ V_{cls}, F^f @ V_{cls})$ where $s \in \mathbb{R}^{J*2}$. A Linear layer is used to project the similarity score to logits along with a softmax operation $l=Softmax(Linear(s))$. We supervise the logits using traditional cross entropy loss $\mathcal{L}_{cls} = CE(l, label)$.

The total loss $\mathcal{L}$ can be formulated in two parts, classification loss $\mathcal{L}_{cls}$ and concept mapping loss $\mathcal{L}_M$:
\begin{align}
    &\mathcal{L} = \mathcal{L}_{cls} + \alpha \mathcal{L}_M.
\end{align}

\begin{table}[t]
\centering
\caption{Stability analysis of Concept Discovery. We use $r=1,2,3$ (r frames per video) across four dataset settings. The results demonstrate the stability of concept discovery process.}
\label{tab:concept_discovery_stability}
\resizebox{0.9\columnwidth}{!}{%
\begin{tabular}{cccccc}
\toprule
\textbf{Dataset}   & \textbf{Class} & \textbf{Top-5} & \textbf{Top-10} & \textbf{Top-15} & \textbf{Random}         \\ \midrule
\multirow{2}{*}{C} & live           & 0.9795         & 0.9413          & 0.8863          & \multirow{9}{*}{0.0552} \\
                   & fake & 0.9617 & 0.9232 & 0.8599 &  \\ \cmidrule{1-5}
\multirow{2}{*}{M} & live & 0.9705 & 0.9302 & 0.8099 &  \\
                   & fake & 0.9660 & 0.9313 & 0.8212 &  \\ \cmidrule{1-5}
\multirow{2}{*}{O} & live & 0.9581 & 0.9116 & 0.8078 &  \\
                   & fake & 0.9887 & 0.9796 & 0.9010 &  \\ \cmidrule{1-5}
\multirow{2}{*}{I} & live & 0.9659 & 0.9308 & 0.8195 &  \\
                   & fake & 0.9773 & 0.9501 & 0.8447 &  \\ \cmidrule{1-5}
Avg.               & All  & 0.9710 & 0.9373 & 0.8438 &  \\ \bottomrule
\end{tabular}%
}
\vspace{-0.3cm}
\end{table}

\section{Visual Concept Discovery Analysis}

In this section, we analyze the automatically discovered visual concepts and the generated heatmaps through Visual Concept-driven Enhancement (VCE) process. We choose to analyze CelebA-Spoof \cite{celebaspoof} dataset which is a large-scale PAD dataset containing rich attributes on face, illumination, environment and attack types.

\begin{figure}[t]
    \centering
    \includegraphics[width=\columnwidth]{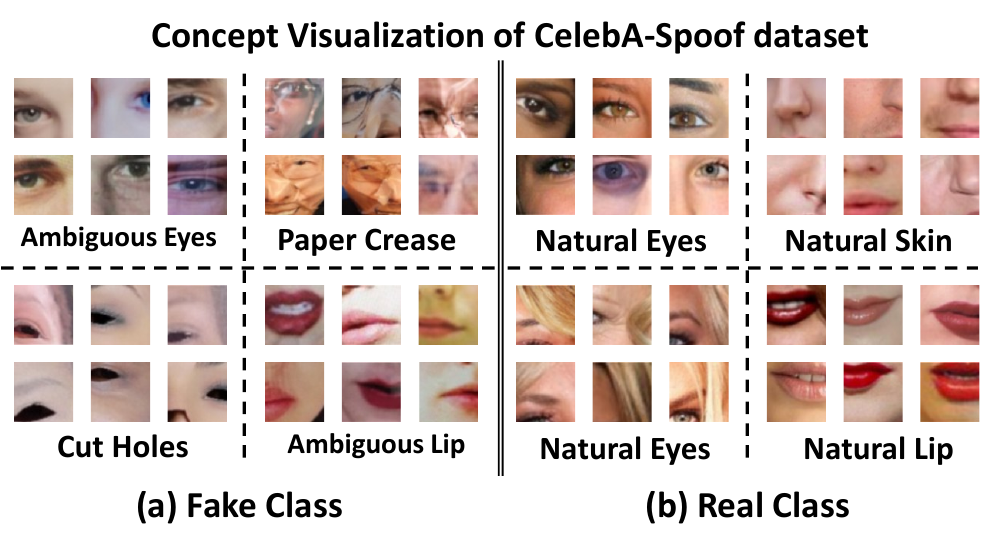}
    \vspace{-0.5cm}
    \caption{Visualization of discovered visual concepts in CelebA-Spoof dataset.}
    \label{fig:concepts_ca}
\end{figure}

\begin{figure}[t]
    \centering
    \includegraphics[width=\columnwidth]{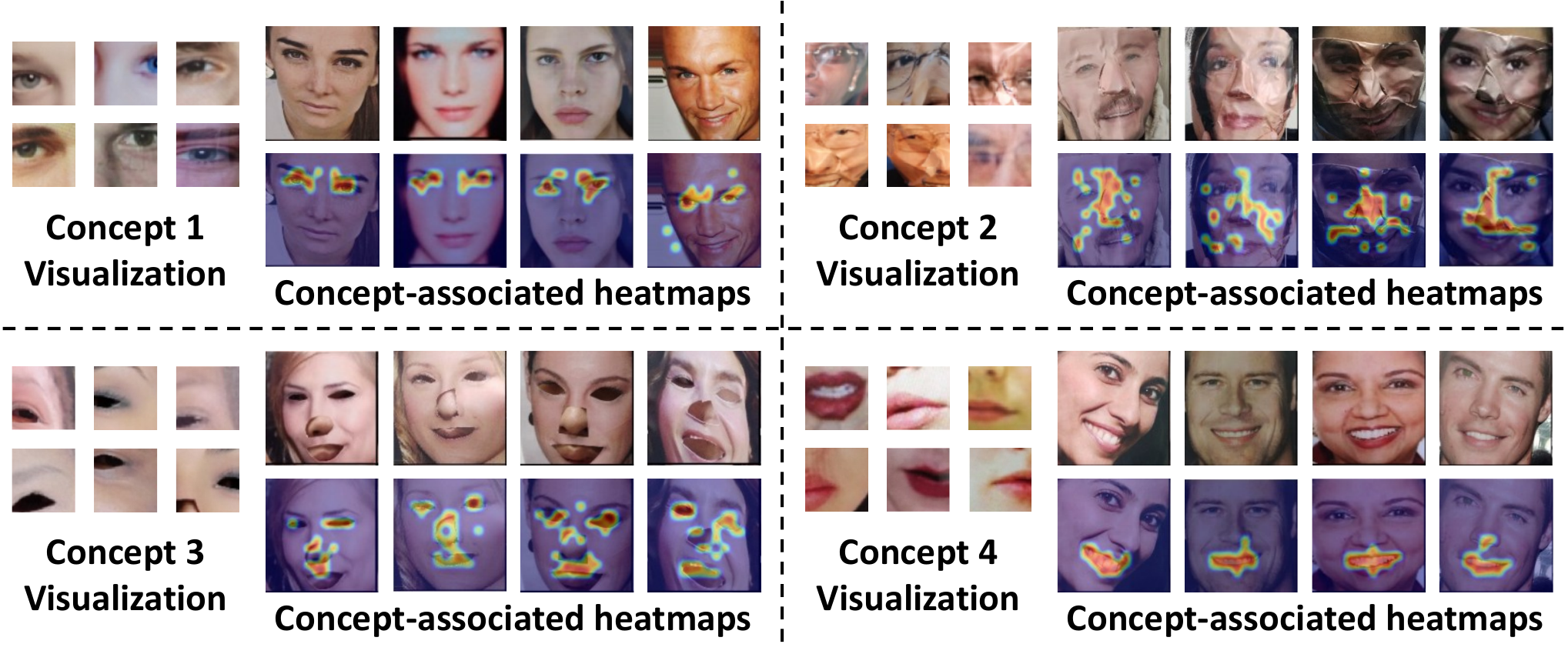}
    \caption{Visualization of concept-associated heatmaps in CelebA-Spoof dataset.}
    \label{fig:heatmaps_ca}
\end{figure}

\subsection{Concept Discovery Stability}

To evaluate the stability of our concept discovery algorithm, we conduct experiments with three different settings ($r=1,2,3$) on each dataset-class combination. For each setting, we repeat the concept discovery process with three different random seeds. For each pair of runs, we compute the cosine similarity between the extracted concept basis vectors and apply maximum bipartite matching (Hungarian algorithm) to find the optimal one-to-one correspondence. Note that Semi-NMF uses a deterministic K-means initialization, and thus different random seeds do not introduce additional variation.

\Cref{tab:concept_discovery_stability} reports the average cosine similarity of the Top-$K$ matched concepts across all pairwise comparisons. The Top-5 concepts achieve 97.1\% average similarity, Top-10 concepts achieve 93.7\%, and all 15 concepts achieve 84.4\%, all significantly higher than the random baseline (5.5\%). We also observe that the most discriminative concepts (e.g., paper edges, cut holes) consistently appear within the Top-10 concepts. These results indicate that the concept discovery process is highly stable across different $r$ values.

\subsection{Visual Concept Visualization}

We employ ActMax \cite{actmax}, a technique used in the XAI domain for concept visualization,  to project the concepts, originally represented as features, into the RGB space. Specifically, ActMax selects the top-k most activated patches for each concept and treats them as representatives.

The discovered concepts in CelebA-Spoof dataset are presented in \Cref{fig:concepts_ca}. In \Cref{fig:concepts_ca}(a), VCE identifies semantic attack concepts, such as the `cut holes' made for the eyes or mouth and the `fold lines' created by bending the paper, whereas others may be more subtle, such as the ambiguity introduced by print attacks. Notably, although some concept visualization may appear similar for real and fake classes, they are distinguishable in the feature space. The results demonstrate that our method can uncover meaningful visual concepts.

\subsection{Concept-Associated Heatmap Visualization}

With the discovered visual concepts, VCE generates concept-associated heatmaps for each sample. After normalization, the heatmaps are resized to 224 × 224 for clearer visualization. The visualizations on the CelebA-Spoof dataset are presented in \Cref{fig:heatmaps_ca}, where red denotes stronger attention and blue denotes weaker attention. As shown, the generated heatmaps align well with the discovered visual concepts. For instance, in \Cref{fig:concepts_ca}, the heatmaps of concept 1 and concept 4 highlight the eyes and mouth, while those of concept 2 and concept 3 capture clear attack cues such as `fold lines' and `cut holes'.

\begin{table*}[t]
\centering
\caption{The results of P1 evaluation on ICMO datasets. Note that * indicates the corresponding method using CelebA-Spoof \cite{celebaspoof} as the supplementary source dataset. $\downarrow$ / $\uparrow$ represents that the smaller/larger value, the better performance. We \textbf{BOLD} the best results and \ul{UNDERLINE} the second best results of CLIP-based methods with and without the CelebA-Spoof dataset.}
\label{tab:p1}
\resizebox{\textwidth}{!}{
\begin{tabular}{rccccccccc}
\toprule
\multirow{2}{*}{Method}                          & \multicolumn{2}{c}{OCI $\rightarrow$ M} & \multicolumn{2}{c}{OMI $\rightarrow$ C} & \multicolumn{2}{c}{OCM $\rightarrow$ I} & \multicolumn{2}{c}{ICM $\rightarrow$ O} & Avg.                  \\ \cmidrule(lr){2-3} \cmidrule(lr){4-5} \cmidrule(lr){6-7} \cmidrule(lr){8-9} \cmidrule(lr){10-10}
& HTER(\%) $\downarrow$  & AUC $\uparrow$ & HTER(\%) $\downarrow$  & AUC $\uparrow$ & HTER(\%) $\downarrow$  & AUC $\uparrow$ & HTER(\%) $\downarrow$  & AUC $\uparrow$ & HTER(\%) $\downarrow$ \\ \midrule
SDA (AAAI'21) \cite{sda}                         & 15.40                  & 91.80          & 24.50                  & 84.40          & 15.60                  & 90.10          & 23.10                  & 84.30          & 19.65                 \\
DRDG (IJCAI'21) \cite{drdg}                      & 12.43                  & 95.81          & 19.05                  & 88.79          & 15.56                  & 91.79          & 16.63                  & 91.75          & 15.66                 \\
FGHV (AAAI'22) \cite{FGHV}                       & 9.17                   & 96.92          & 12.47                  & 93.47          & 16.29                  & 90.11          & 13.58                  & 93.55          & 12.88                 \\
GDA (ECCV'22) \cite{GDA}                         & 9.20                   & 98.00          & 12.20                  & 93.00          & 10.00                  & 96.00          & 14.40                  & 92.60          & 11.45                 \\
SSAN-R (CVPR'22) \cite{ssanr}                    & 6.67                   & 98.75          & 10.00                  & 96.67          & 8.88                   & 96.79          & 13.72                  & 93.63          & 9.82                  \\
SA-FAS (CVPR'23) \cite{safas}                    & 5.95                   & 96.55          & 8.78                   & 95.37          & 6.58                   & 97.54          & 10.00                  & 96.23          & 7.82                  \\
IADG (CVPR'23) \cite{iadg}                       & 5.41                   & 98.19          & 8.70                   & 96.40          & 10.62                  & 94.50          & 8.86                   & 97.14          & 8.40                  \\
UDG-FAS (ICCV'23) \cite{udgfas}                  & 5.95                   & 98.47          & 9.82                   & 96.76          & 5.86                   & 98.62          & 10.97                  & 95.36          & 8.15                  \\
DiVT-M (WACV'23) \cite{divtm}                    & 2.86                   & 99.14          & 8.67                   & 96.92          & 3.71                   & 99.29          & 13.06                  & 94.04          & 7.08                  \\
TTDG (CVPR'24) \cite{ttdg}                       & 4.16                   & 98.48          & 7.59                   & 98.18          & 9.62                   & 98.18          & 10.00                  & 96.15          & 7.84                  \\
GAC-FAS (CVPR'24) \cite{gacfas}                  & 5.00                   & 97.56          & 8.20                   & 95.16          & 4.29                   & 98.87          & 8.60                   & 97.16          & 6.52                  \\ \midrule
CLIP (ICML'21) \cite{clip}                       & 4.04                   & 99.13          & 5.00                   & 98.89          & 6.57                   & 98.45          & 6.09                   & 98.12          & 5.43                  \\
CoOp (IJCV'22) \cite{coop}                       & 3.86                   & 99.08          & \ul{2.33}             & 98.92          & 6.07                   & 98.52          & 5.83                   & 98.97          & 4.37                  \\
CoCoOp (CVPR'22) \cite{cocoop}                   & 4.16                   & 99.01          & 5.17                   & 98.19          & 6.21                   & 98.50          & 6.00                   & 98.49          & 5.39                  \\
CFPL-FAS (CVPR'24) \cite{cfpl}                   & 3.09                   & \textbf{99.45} & 2.56                   & \ul{99.10}    & \ul{5.43}             & 98.41          & \ul{3.33}             & \ul{99.05}    & 3.60                  \\
S-CPTL (ACM MM'24) \cite{scptl}                  & \textbf{1.43}          & 99.17          & \textbf{0.89}          & 99.00          & 6.86                   & \ul{98.63}    & 4.12                   & 99.02          & \ul{3.33}            \\
CCPE (TMM'25) \cite{ccpe}                  & 3.10          & 99.21          & 1.33          & \textbf{99.36}          & 6.08                   & 94.36    & 5.57                   & 98.49          & 4.02            \\
\rowcolor[HTML]{C0C0C0} \textbf{CPG-PAD (Ours)}  & \ul{2.86}             & \ul{99.41}    & 2.67                   & 99.27 & \textbf{1.29}          & \textbf{99.90} & \textbf{1.49}          & \textbf{99.80} & \textbf{2.08}         \\ \midrule
ViT* (ECCV'22) \cite{vitfas}                     & 1.58                   & \ul{99.68}    & 5.70                   & 98.91          & 9.25                   & 97.15          & 7.47                   & 98.42          & 6.00                  \\
FLIP-MCL* (ICCV'23) \cite{flip}                  & 4.95                   & 98.11          & \ul{0.54}             & \textbf{99.98} & 4.25                   & 99.07          & 2.31                   & 99.63    & 3.01                  \\
CFPL-FAS* (CVPR'24) \cite{cfpl}                  & \ul{1.43}             & 99.28          & 2.56                   & 99.10          & 5.43                   & 98.41          & 2.50                   & 99.42          & 2.98                  \\
FGPL* (ACM MM'24) \cite{fgpl}                    & 2.86                   & 98.12          & 3.89                   & 98.19          & \ul{3.50}             & \ul{99.54}    & \ul{1.77}             & 99.23          & 3.01                  \\
S-CPTL* (ACM MM'24) \cite{scptl}                 & \textbf{1.25}          & 99.35          & \textbf{0.52}          & \ul{99.90}    & 5.84                   & 98.78          & 3.95                   & \textbf{99.72} & \ul{2.89}            \\
CCPE* (TMM'25) \cite{ccpe}                  & 2.86          & 99.24          & 1.30          & \textbf{99.98}          & 4.15                   & 99.31    & 3.64                   & \ul{99.67}          & 2.99            \\
\rowcolor[HTML]{C0C0C0} \textbf{CPG-PAD* (Ours)} & 1.67                   & \textbf{99.74} & 0.78                   & 99.86          & \textbf{0.71}          & \textbf{99.80} & \textbf{1.58}          & 99.63    & \textbf{1.19}         \\  \bottomrule
\end{tabular}
}
\end{table*}

\begin{table*}[t]
\centering
\caption{The results of P1 evaluation on ICMO datasets. Note that * indicates using CelebA-Spoof \cite{celebaspoof} as the supplementary source dataset and \dag \ indicates methods using LLMs. $\downarrow$ / $\uparrow$ represents that the smaller/larger value, the better performance. We \textbf{BOLD} the best average HTER(\%).}
\label{tab:p1_extra}
\resizebox{\textwidth}{!}{
\begin{tabular}{rccccccccc}
\toprule
\multirow{2}{*}{Method}                          & \multicolumn{2}{c}{OCI $\rightarrow$ M} & \multicolumn{2}{c}{OMI $\rightarrow$ C} & \multicolumn{2}{c}{OCM $\rightarrow$ I} & \multicolumn{2}{c}{ICM $\rightarrow$ O} & Avg.                  \\ \cmidrule(lr){2-3} \cmidrule(lr){4-5} \cmidrule(lr){6-7} \cmidrule(lr){8-9} \cmidrule(lr){10-10}
& HTER(\%) $\downarrow$  & AUC $\uparrow$ & HTER(\%) $\downarrow$  & AUC $\uparrow$ & HTER(\%) $\downarrow$  & AUC $\uparrow$ & HTER(\%) $\downarrow$  & AUC $\uparrow$ & HTER(\%) $\downarrow$ \\ \midrule
TF-FAS\dag (ECCV'24) \cite{tffas} &	3.44 &	99.42 &	0.81 &	99.92 &	2.24 &	99.67 &	2.26 &	99.48 &	2.19 \\
\rowcolor[HTML]{C0C0C0} \textbf{CPG-PAD (Ours)}  & 2.86          & 99.41    & 2.67                   & 99.27 & 1.29          & 99.90 & 1.49          & 99.80 & \textbf{2.08}         \\ \midrule
TF-FAS*\dag (ECCV'24) \cite{tffas} &	1.49 &	99.80 &	0.58 &	99.99 &	1.56 &	99.89 &	1.43 &	99.93 &	1.27 \\
I-FAS*\dag (AAAI'25) \cite{ifas} &	0.32 &	99.88 &	0.04 &	99.99 &	3.22 &	98.48 &	1.74 &	99.66 &	1.33 \\
\rowcolor[HTML]{C0C0C0} \textbf{CPG-PAD* (Ours)} & 1.67                   & 99.74 & 0.78                   & 99.86          & 0.71          & 99.80 & 1.58          & 99.63    & \textbf{1.19}         \\ \bottomrule
\end{tabular}
}
\vspace{-0.3cm}
\end{table*}

\begin{table}[t]
\centering
\caption{Extra metric result of P1 evaluation. We report BPCER10, BPCER20, BPCER100 metric for fair comparison following standard ISO/IEC 30107-3 \cite{ISO-IEC-30107-3-PAD-metrics-2023}. We also report results using EER threshold on validation set (@val-thr).}
\label{tab:p1_extra2}
\resizebox{\columnwidth}{!}{
\begin{tabular}{r|r|ccccc}
\toprule
Method & Metric & OCI $\rightarrow$ M & OMI $\rightarrow$ C & OCM $\rightarrow$ I & ICM $\rightarrow$ O & Avg. \\ \midrule
\multirow{6}{*}{FLIP \cite{flip}} & APCER@val-thr & 7.14 & 6.00 & 1.43 & 8.75 & 5.83 \\
& BPCER@val-thr  & 4.29 & 0.22 & 6.00 & 1.01 & 2.88 \\
& HTER@val-thr  & 5.71 & 3.11 & 3.71 & 4.88 & 4.35 \\
& BPCER10  & 3.81 & 0.22 & 2.14 & 0.80 & 1.74 \\
& BPCER20  & 5.24 & 0.22 & 2.86 & 1.98 & 2.57 \\
& BPCER100 & 20.48 & 0.67 & 6.00 & 4.69 & 7.96 \\ \midrule
\multirow{6}{*}{CPG-PAD} & APCER@val-thr & 1.90 & 2.00 & 3.00 & 1.64 & 2.14 \\
& BPCER@val-thr  & 2.86 & 5.33 & 0.00 & 1.46 & 2.41 \\
& HTER@val-thr  & 2.38 & 3.67 & 1.50 & 1.55 & 2.28 \\
& BPCER10 & 1.43 & 1.33 & 0.00 & 0.00 & 0.69 \\
& BPCER20 & 1.43 & 2.00 & 0.00 & 0.49 & 0.98 \\
& BPCER100 & 5.71 & 8.00 & 1.43 & 4.88 & 5.01 \\ \bottomrule
\end{tabular}%
}
\vspace{-0.5cm}
\end{table}

\section{Experiments}

\subsection{Experimental Setups}

\subsubsection{Datasets}

{We employ nine datasets to evaluate the performance of our method, including MSU-MFSD (M) \cite{msumfsd}, CASIA-FASD (C) \cite{casiafasd}, Idiap Replay-Attack (I) \cite{replayattack}, OULU-NPU (O) \cite{oulunpu}, CASIA-SURF (S) \cite{casiasurf}, CASIA-CeFA (C) \cite{casiacefa}, WMCA (W) \cite{wmca}, CelebA-Spoof \cite{celebaspoof} and SiW-Mv2 \cite{siwmv2}. Idiap Replay-Attack, CASIA-FASD, MSU-MFSD and OULU-NPU datasets (ICMO) differ in material, lighting, background, and resolution. CASIA-SURF, CASIA-CeFA, and WMCA datasets (CSW) encompass a broader range of topics, various types of attacks, and diverse sampling environments. In all experimental protocols, each dataset is treated as an individual domain, and we use the notation $ XY \rightarrow Z $ to indicate that datasets X and Y serve as the source domain for training, while dataset Z is used as an unseen target domain for evaluation.}

\begin{table*}[t]
\centering
\caption{The results of \textbf{P2} evaluation on CSW datasets. Note that * indicates the corresponding method using CelebA-Spoof \cite{celebaspoof} as the supplementary source dataset. $\downarrow$ / $\uparrow$ represents that the smaller/larger value, the better performance. We \textbf{BOLD} the best results and \ul{UNDERLINED} the second best results of CLIP-based methods with and without the CelebA-Spoof dataset.}
\label{tab:p2}
\resizebox{\textwidth}{!}{
\begin{tabular}{rcccccccccc}
\toprule
\multirow{2}{*}{Methods}                         & \multicolumn{3}{c}{CS $\rightarrow$ W}                                                          & \multicolumn{3}{c}{SW $\rightarrow$ C}                                                          & \multicolumn{3}{c}{CW $\rightarrow$ S}                                                          & Avg.                  \\ \cmidrule(lr){2-4} \cmidrule(lr){5-7} \cmidrule(lr){8-10} \cmidrule(lr){11-11}
& HTER(\%) $\downarrow$ & AUC $\uparrow$ & \begin{tabular}[c]{@{}c@{}}TPR@\\ FPR=1\%\end{tabular} & HTER(\%) $\downarrow$ & AUC $\uparrow$ & \begin{tabular}[c]{@{}c@{}}TPR@\\ FPR=1\%\end{tabular} & HTER(\%) $\downarrow$ & AUC $\uparrow$ & \begin{tabular}[c]{@{}c@{}}TPR@\\ FPR=1\%\end{tabular} & HTER(\%) $\downarrow$ \\  \midrule
ViT (ECCV'22) \cite{vitfas}                      & 21.04                 & 89.12          & 30.09                                                  & 17.12                 & 89.05          & \textbf{22.71}                                         & 17.16                 & 90.25          & 30.23                                                  & 18.44                 \\
CLIP-V (ICML'21) \cite{clip}                     & 20.00                 & 87.72          & 16.44                                                  & 17.67                 & 89.67          & \ul{20.70}                                            & \ul{8.32}            & \ul{97.23}    & 57.28                                                  & 15.33                 \\
CLIP (ICML'21) \cite{clip}                       & 17.05                 & 89.37          & 8.17                                                   & 15.22                 & \ul{91.99}    & 17.08                                                  & 9.34                  & 96.62          & \ul{60.75}                                            & 13.87                 \\
CoOp (IJCV'22) \cite{coop}                       & 9.52                  & 90.49          & 10.68                                                  & 18.30                 & 87.47          & 11.50                                                  & 11.37                 & 95.46          & 40.40                                                  & 13.06                 \\
CoCoOp (CVPR'22) \cite{cocoop}                   & 13.89                 & 90.74          & -                                                      & 15.49                 & 89.40          & -                                                      & 13.76                 & 95.59          & -                                                      & 14.38                 \\
S-CPTL (ACM MM'24) \cite{scptl}                  & \ul{8.99}            & 94.01          & -                                                      & \textbf{12.78}        & 91.64          & -                                                      & 9.48                  & 95.83          & -                                                      & \ul{10.42}           \\
FGPL (ACM MM'24) \cite{fgpl}                     & 14.05                 & 92.65          & \ul{33.33}                                            & 19.00                 & 88.53          & 13.33                                                  & 11.00                 & 94.72          & 34.00                                                  & 14.68                 \\
CFPL-FAS (CVPR'24) \cite{cfpl}                       & 9.04                  & \ul{96.48}    & 25.84                                                  & 14.83                 & 90.36          & 8.33                                                   & 8.77                  & 96.83          & 53.34                                                  & 10.88                 \\
\rowcolor[HTML]{C0C0C0} \textbf{CPG-PAD (Ours)}  & \textbf{8.65}         & \textbf{97.77} & \textbf{67.42}                                         & \ul{13.37}           & \textbf{93.76} & 12.23                                                  & \textbf{7.71}         & \textbf{98.09} & \textbf{71.01}                                         & \textbf{9.91}         \\ \midrule
ViT* (ECCV'22) \cite{vitfas}                     & 7.98                  & 97.97          & 73.61                                                  & 11.13                 & 95.46          & 47.59                                                  & 13.35                 & 94.13          & 49.97                                                  & 10.82                 \\
FLIP-MCL* (ICCV'23) \cite{flip}                  & 4.46                  & \ul{99.16}    & \ul{83.86}                                            & 9.66                  & 96.69          & 59.00                                                  & 11.71                 & 95.21          & \ul{57.98}                                            & 8.61                  \\
S-CPTL* (ACM MM'24) \cite{scptl}                 & \textbf{4.22}         & \textbf{99.30} & -                                                      & 9.59                  & \ul{96.73}    & -                                                      & 10.97                 & \textbf{97.40} & -                                                      & 8.33                  \\
CFPL-FAS* (CVPR'24) \cite{cfpl}                      & \ul{4.40}            & 99.11          & \textbf{85.23}                                         & \ul{8.13}            & 96.70          & \ul{62.41}                                            & \textbf{8.50}         & 97.00          & 55.66                                                  & \ul{7.01}            \\
\rowcolor[HTML]{C0C0C0} \textbf{CPG-PAD* (Ours)} & 6.99                  & 98.20          & 76.48                                                  & \textbf{4.19}         & \textbf{99.16} & \textbf{74.09}                                         & \ul{8.59}            & \ul{97.37}    & \textbf{61.15}                                         & \textbf{6.59}         \\ \bottomrule
\end{tabular}
}
\end{table*}

\begin{table*}[htbp]
	\centering
	\caption{The results of \textbf{P4} evaluation. We report the HTER(\%) results on the different 12 scenarios on ICMO datasets. We only \textbf{BOLD} the best results and Avg. denotes the mean HTER(\%).}
	\label{tab:p4}
	\resizebox{\textwidth}{!}{%
\begin{tabular}{rccccccccccccc}
\toprule
Methods                                          & C → I         & C → M         & C → O         & I → C         & I → M         & I → O         & M → C         & M → I         & M → O         & O → C         & O → I         & O → M         & Avg.          \\ \midrule
ADDA (CVPR'17) \cite{adda}                       & 41.8          & 36.6          & -             & 49.8          & 35.1          & -             & 39.0          & 35.2          & -             & -             & -             & -             & 39.6          \\
DRCN (ECCV'16) \cite{drcn}                       & 44.4          & 27.6          & -             & 48.9          & 42.0          & -             & 28.9          & 36.8          & -             & -             & -             & -             & 38.1          \\
DupGAN (CVPR'18) \cite{dupgan}                   & 42.4          & 33.4          & -             & 46.5          & 36.2          & -             & 27.1          & 35.4          & -             & -             & -             & -             & 36.8          \\
KSA (TIFS'18) \cite{ksa}                         & 39.3          & 15.1          & -             & 12.3          & 33.3          & -             & 9.1           & 34.9          & -             & -             & -             & -             & 24.0          \\
DR-UDA (TIFS'20) \cite{druda}                    & 15.6          & 9.0           & 28.7          & 34.2          & 29.0          & 38.5          & 16.8          & 3.0           & 30.2          & 19.5          & 25.4          & 27.4          & 23.1          \\
USDAN-Un (PR'21) \cite{usdan}                    & 16.0          & 9.2           & -             & 30.2          & 25.8          & -             & 13.3          & 3.4           & -             & -             & -             & -             & 16.3          \\
GDA (ECCV'22) \cite{GDA}                         & 15.1          & 5.8           & -             & 29.7          & 20.8          & -             & 12.2          & 2.5           & -             & -             & -             & -             & 14.4          \\
CDFTN-L (AAAI'23) \cite{cdftn}                   & \textbf{1.7}  & 8.1           & 29.9          & \textbf{11.9} & 9.6           & 29.9          & 8.8           & \textbf{1.3}  & 25.6          & 19.1          & 5.8           & 6.3           & 13.2          \\
\rowcolor[HTML]{C0C0C0} \textbf{CPG-PAD (Ours)}  & 7.14          & \textbf{4.29} & \textbf{2.99} & 14.67         & \textbf{7.38} & \textbf{6.61} & \textbf{4.00} & 2.29          & \textbf{2.48} & \textbf{4.78} & \textbf{5.71} & \textbf{1.67} & \textbf{5.33} \\ \midrule
FLIP-MCL* (ICCV'23) \cite{flip}                  & 10.57         & \textbf{7.15} & 3.19          & \textbf{0.68} & 7.22          & 4.22          & \textbf{0.19} & 5.88          & 3.95          & \textbf{0.19} & 5.69          & 8.40          & 4.84          \\
\rowcolor[HTML]{C0C0C0} \textbf{CPG-PAD* (Ours)} & \textbf{4.29} & 8.33          & \textbf{3.02} & 3.33          & \textbf{5.71} & \textbf{3.88} & 0.67          & \textbf{0.71} & \textbf{2.68} & 1.22          & \textbf{1.00} & \textbf{5.48} & \textbf{3.36} \\ \bottomrule
\end{tabular}
}
\end{table*}

\subsubsection{Protocols}

Following previous works \cite{cfpl, flip}, we adopt 4 evaluation protocols to assess the generalizability of our method: multi-source evaluation (\textbf{P1}), limited-source evaluation (\textbf{P2} and \textbf{P3}), single-source evaluation (\textbf{P4}), unknown PAI (Presentation Attack Instruments) evaluation (\textbf{P5}). For each protocol, we train the model on the training sets of the source domains and evaluate it on the test set of the target domain. Following previous works, evaluation is conducted at the video level, where two frames are sampled from each video and their scores are averaged to obtain the final video-level prediction.

\subsubsection{Evaluation Metrics}

Following standard evaluation principles, we employed three key metrics, HTER, AUC, and TPR, to assess the model’s performance. (1) HTER (Half Total Error Rate) quantifies the trade-off between false rejection and false acceptance errors, computed as the average of the False Rejection Rate (FRR) and the False Acceptance Rate (FAR). (2) AUC (Area Under the Curve) measures the overall classification performance by representing the area under the Receiver Operating Characteristic (ROC) curve. (3) TPR (True Positive Rate) evaluates the model’s ability to correctly identify attack samples, with the classification threshold adjustable based on specific application requirements to optimize performance. Besides, following standard ISO/IEC 30107-3 \cite{ISO-IEC-30107-3-PAD-metrics-2023} for biometric PAD, we report (1) the BPCERs (Bona Fide Presentation Classification Error Rate) observed at APCER (Attack Presentation Classification Error Rate) values or security thresholds of 1\% (BPCER100), 5\% (BPCER20), and 10\% (BPCER10) and draw DET curves for \textbf{P1} protocol. 

\textbf{Thresholding protocol and limitation:}
Following common cross-database protocols, HTER is computed using the EER threshold estimated from the target test scores for direct comparison with prior works. We acknowledge that such a posteriori thresholding may introduce optimistic bias, and therefore additionally report results under a stricter setting where the threshold is determined on the validation set and then evaluated on the test set (\textbf{P1}). We also provide BPCER/APCER metrics and DET curves to complement HTER.

\subsubsection{Implementation Details}

For the PAD task, we use a pretrained CLIP~\cite{clip} model from OpenAI\footnote{https://github.com/OpenAI/CLIP} with ViT-B/16 as the image encoder, the same as previous methods. We keep the parameters in the text encoder unchanged, and finetune the image encoder with a simple Convpass adapter. We resize the image to 224×224 with a batch size of 24, and train all models for 50 epochs. We use the Adam optimizer and set the learning rate to 1e-4 for training. In all experiments we set $r=2$, $K=15$ and set the number of learnable concept-informed prompts $J=K=15$. We set the depth of VPD $Depth=4$, the coefficient of concept mapping loss $\alpha=0.01$.

\begin{table*}[t]
\centering
\caption{The results of P5 evaluation on SiW-Mv2 dataset. We follow \cite{foundation_model_pad} and use leave-one-out strategy to evaluate the generalizability of unknown PAIs. The results of ViT-B/16 come from \cite{foundation_model_pad}.}
\label{tab:pai}
\resizebox{\textwidth}{!}{%
\begin{tabular}{r|r|cccc|ccc|ccccc|cc|c}
\toprule
\multirow{2}{*}{Approaches} & \multirow{2}{*}{Metrics} & \multicolumn{4}{c|}{Covering} & \multicolumn{3}{c|}{Make-up} & \multicolumn{5}{c|}{3D Attack} & \multicolumn{2}{c|}{2D Attack} & \multirow{2}{*}{Avg.$\pm$Std.} \\ 
 &  & FunE. & PEye & PMouth & PaperG. & Ob. & Impers. & Cosmetic & HalfM. & Silicone & TransM. & Paper & Mann. & Replay & Print &  \\ \midrule
\multirow{2}{*}{SiW-Mv2 baseline\cite{siwmv2}} & HTER(\%) $\downarrow$ & 29.50 & 2.70 & 1.10 & 11.90  & 1.30  & 24.50 & 10.90 & 8.00  & 9.20  & 0.00  & 0.60  & 4.00 & 17.90 & 9.60  & 9.40$\pm$8.80 \\
& BPCER100(\%) $\downarrow$ &  91.10 & 63.00 & 11.60 & 96.00 & 1.70 & 76.20 & 60.80 & 38.60 & 52.50 & 0.00 & 0.00 & 33.4 & 60.70 & 21.10 & 43.34$\pm$33.19 \\ \midrule
\multirow{4}{*}{ViT-B/16} & HTER(\%) $\downarrow$ & 13.46 & 0.19 & 0.39  & 1.62  & 8.60  & 0.19  & 11.37 & 3.51  & 0.58  & 3.21  & 0.19  & 0.19 & 16.27 & 11.72 & 5.11$\pm$5.87 \\
& BPCER10(\%) $\downarrow$ & 14.67 & 0.39 & 0.39  & 0.39  & 8.11  & 0.39  & 15.44 & 2.70  & 0.39  & 2.32  & 0.00  & 0.39 & 17.76 & 15.44 & 5.63$\pm$7.04   \\
& BPCER20(\%) $\downarrow$ & 33.20 & 0.39 & 0.77  & 1.16  &22.01  & 0.39  & 20.46 & 4.25  & 0.77  & 3.09  & 0.39  & 0.39 & 23.55 & 26.25 & 9.79$\pm$12.21   \\
& BPCER100(\%) $\downarrow$ & 50.19 & 0.39 & 0.77  & 1.93  &22.39  & 0.39  & 26.25 & 8.49  & 1.16  & 7.34  & 0.39  & 0.39 & 32.05 & 42.47 & 13.90$\pm$17.47   \\ \midrule
\multirow{4}{*}{CLIP} & HTER(\%) $\downarrow$ & 10.64 & 0.19 & 0.19 & 0.38 & 14.79 & 0.19 & 8.19 & 0.19 & 0.19 & 2.77 & 0.00 & 0.19 & 2.62 & 1.85 & 3.03$\pm$4.54 \\
& BPCER10(\%) $\downarrow$ & 11.11 & 0.38 & 0.38 & 0.38 & 9.58 & 0.38 & 6.90 & 0.38 & 0.38 & 1.53 & 0.00 & 0.38 & 2.30 & 0.77 & 2.49$\pm$3.64 \\
& BPCER20(\%) $\downarrow$ & 27.59 & 0.38 & 0.38 & 0.38 & 10.34 & 0.38 & 9.58 & 0.38 & 0.38 & 2.68 & 0.00 & 0.38 & 2.30 & 1.15 & 4.02$\pm$7.31 \\
& BPCER100(\%) $\downarrow$ & 63.98 & 0.38 & 0.38 & 0.77 & 10.34 & 0.38 & 61.69 & 0.38 & 0.38 & 4.60 & 0.00 & 0.38 & 6.13 & 5.36 & 11.08$\pm$21.34 \\ \midrule
\multirow{4}{*}{CPG-PAD} & HTER(\%) $\downarrow$ & 4.45 & 0.19 & 0.19 & 0.19 & 0.77 & 0.19 & 4.19 & 0.19 & 0.19 & 0.57 & 0.00 & 0.19 & 2.62 & 0.38 & \textbf{1.02$\pm$1.49} \\
& BPCER10(\%) $\downarrow$ & 2.30 & 0.38 & 0.38 & 0.38 & 1.53 & 0.38 & 0.77 & 0.38 & 0.38 & 0.38 & 0.00 & 0.38 & 1.92 & 0.77 & \textbf{0.74$\pm$0.66} \\
& BPCER20(\%) $\downarrow$ & 3.83 & 0.38 & 0.38 & 0.38 & 1.53 & 0.38 & 2.68 & 0.38 & 0.38 & 0.77 & 0.00 & 0.38 & 2.68 & 0.77 & \textbf{1.07$\pm$1.12} \\
& BPCER100(\%) $\downarrow$ & 57.47 & 0.38 & 0.38 & 0.38 & 1.53 & 0.38 & 22.61 & 0.38 & 0.38 & 1.15 & 0.00 & 0.38 & 6.90 & 0.77 & \textbf{6.65$\pm$15.23} \\ \bottomrule
\end{tabular}%
}
\vspace{-0.3cm}
\end{table*}

\begin{table}[t]
\centering
\caption{The results of \textbf{P3} evaluation. We \textbf{BOLD} the best results.}
\label{tab:p3}
\resizebox{\columnwidth}{!}{
\begin{tabular}{rccccc}
\toprule
& \multicolumn{2}{c}{MI $\rightarrow$ C} & \multicolumn{2}{c}{MI $\rightarrow$ O} & Avg.          \\ \cmidrule{2-3} \cmidrule{4-5} \cmidrule{6-6}
\multirow{-2}{*}{Methods} & HTER         & AUC               & HTER          & AUC               & HTER     \\ \midrule
SSDG-R (CVPR'20) \cite{ssdgr}         & 19.86             & 86.46              & 27.92             & 78.72              & 23.89         \\
SSAN-R (CVPR'22) \cite{ssanr}         & 25.56             & 83.89              & 24.44             & 82.86              & 25.00         \\
HFN+MP (TIFS'22) \cite{hfnmp}         & 30.89             & 72.48              & 20.94             & 85.71              & 25.92         \\
CIFAS (ICME'22)  \cite{cifas}         & 22.67             & 83.89              & 24.63             & 81.48              & 23.65         \\
DiVT-M (WACV'23) \cite{divtm}         & 20.11             & 86.71              & 23.61             & 85.73              & 21.86         \\
DGUA-FAS (ICIP'23) \cite{dguafas}        & 19.22             & 86.81              & 20.05             & 88.75              & 19.64         \\
BUDoPT (ECCV'24) \cite{budopt}         & 5.33              & 98.92              & 5.94              & 98.37              & 5.64          \\
\rowcolor[HTML]{C0C0C0} 
\textbf{CPG-PAD (Ours)}               & \textbf{2.67}     & \textbf{99.25}     & \textbf{2.25}     & \textbf{99.81}     & \textbf{2.46} \\ \bottomrule
\end{tabular}%
}
\vspace{-0.5cm}
\end{table}

\subsection{PAD Performance}

\subsubsection{Multi-source evaluation (P1)}

\textbf{Multi-source evaluation (P1).} In \textbf{P1} protocol, we construct a domain generalization benchmark using four datasets, ICMO, resulting in 4 scenarios where three datasets are seen as source domains and the remaining one serves as the target domain. We include traditional PAD methods \cite{sda, drdg, FGHV, GDA, ssanr, safas, iadg, udgfas, divtm, ttdg, gacfas} and current proposed CLIP-based methods \cite{clip, coop, cocoop, cfpl, scptl, ccpe} as baseline. As shown in \Cref{tab:p1}, we can draw the following conclusions: (1) CLIP-based methods outperform the traditional DG methods, even the finetuned CLIP model has a lower average HTER (5.43\%) compared to the most competitive traditional DG method GAC-FAS (6.52\%), supporting the motivation that general knowledge from CLIP model can benefit PAD task. (2) Without supplementary training data, the proposed CPG-PAD outperforms SOTA methods with an average HTER of 2.08\%, surpassing S-CPTL \cite{scptl} by 1.25\% and even surpassing S-CPTL* which is trained with additional CelebA-Spoof dataset. On all 8 settings, CPG-PAD achieves top-two performance on 7 of the settings. (3) With supplementary training data (CelebA-Spoof), CPG-PAD also surpasses the recently proposed methods with an average HTER of 1.19\%. The above results and conclusions prove that CPG-PAD can better leverage the general knowledge from pretrained CLIP model and effectively enhance PAD generalizability in multi-source scenarios.

Recent MLLM-based FAS methods \cite{tffas, ifas, tarfas} introduce large language models or multimodal reasoning into PAD, but usually require substantially larger model capacity than CLIP-based frameworks. As we can see in \Cref{tab:p1_extra}, our method CPG-PAD surpasses TF-FAS \cite{tffas} and I-FAS \cite{ifas} on average HTER metric with a small margin. However, achieving a performance level that is nearly on par is already sufficient to highlight the effectiveness of CPG-PAD since CPG-PAD is a vision-only model that does not rely on any extra annotations or LLMs.

\begin{figure}[t]
\centering
\includegraphics[width=0.9\columnwidth]{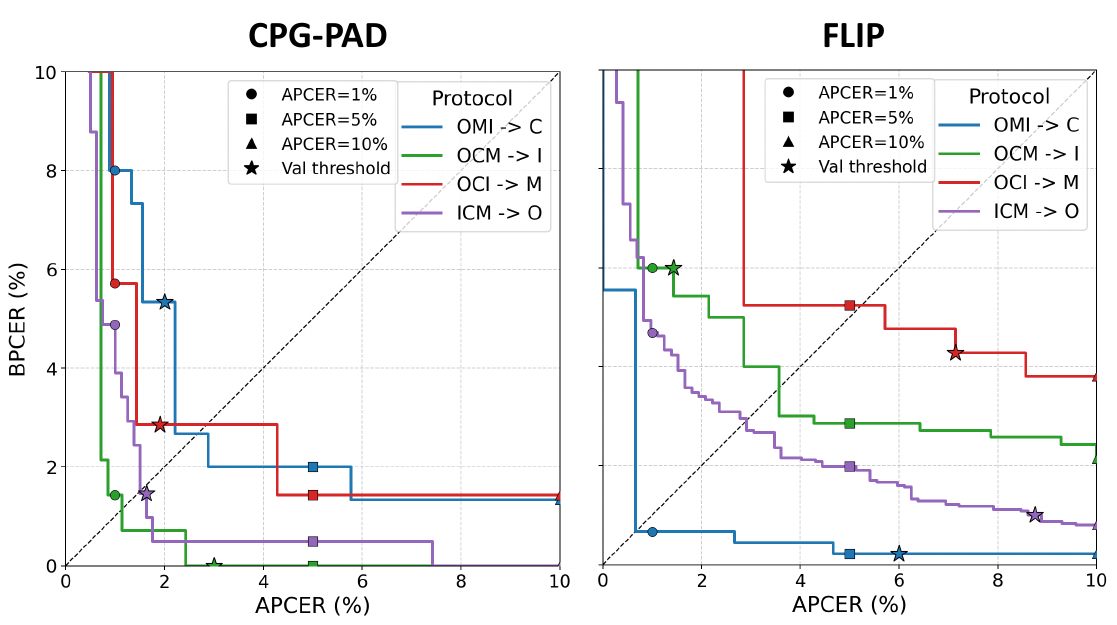}
\caption{DET curve comparison of P1 evaluation.}
\label{fig:det}
\vspace{-0.3cm}
\end{figure}

Following the standard ISO/IEC 30107-3 \cite{ISO-IEC-30107-3-PAD-metrics-2023}, we further compare CPG-PAD with the reproducible FLIP baseline using extra metrics as reported in \Cref{tab:p1_extra2}, together with the DET curves in \Cref{fig:det}. Extra metrics include BPCER10, BPCER20, BPCER100 and BPCER/APCER/HTER results using EER threshold on validation set. Although FLIP achieves better performance in one protocol, CPG-PAD obtains better results in most cases and shows stronger overall operating-point performance. The DET curves further demonstrate that CPG-PAD yields a more favorable APCER-BPCER trade-off.

\subsubsection{Limited-source evaluation (P2 and P3)}

In this evaluation experiment, we test on two protocols (\textbf{P2} and \textbf{P3}) with two datasets as source domains and one dataset as target domain. For \textbf{P2}, we establish a domain generalization benchmark focused on multimodal sensors, incorporating the CSW datasets, leading to 3 sub-experiments where two datasets are seen as source domain and the remaining one serves as target domain. As \Cref{tab:p2} shows, without additional training data, the proposed CPG-PAD outperforms the previous SOTA method S-CPTL \cite{scptl} with an average HTER of 9.91\% and achieves top-two performance across eight of nine metrics. With CelebA-Spoof as additional training data, the proposed CPG-PAD also achieves SOTA performance with an average HTER of 6.59\%. Notably, as the results show, the SW $\rightarrow$ C scenario is difficult for all previous methods since the domain gap is larger than the other two scenarios. However, CPG-PAD* achieves nearly half the HTER of the second-best CFPL-FAS*. For \textbf{P3}, we follow \cite{budopt} and construct 2 scenarios with ICMO datasets which set M and I as source domains and test on C and O separately. In all settings, the proposed CPG-PAD surpasses previous methods, demonstrating its strong generalizability in limited-source PAD scenarios.

\begin{table*}[t]
\centering
\caption{The effect of MLP, CML and VPD. {\color[HTML]{009901} Improvement}/{\color[HTML]{FE0000} Degradation} indicates the comparison against the baseline CLIP \cite{clip}. ``MLP'' indicates Multiple Learnable Prompts, ``CML'' indicates Concept Mapping Loss, ``VPD'' indicates Visual-Prompt Decoder.}
\label{tab:ablation1}
\resizebox{\textwidth}{!}{
\begin{tabular}{ccccccccccccc}
\toprule
\multicolumn{4}{c}{Components}                                                                                                            & \multicolumn{2}{c}{OCI $\rightarrow$ M}                                  & \multicolumn{2}{c}{OMI $\rightarrow$ C}                                  & \multicolumn{2}{c}{OCM $\rightarrow$ I}                                  & \multicolumn{2}{c}{ICM $\rightarrow$ O}                                  & Avg.                                \\ \cmidrule(lr){5-6} \cmidrule(lr){7-8} \cmidrule(lr){9-10} \cmidrule(lr){11-12} \cmidrule(lr){13-13}
CLIP                             & MLP                              & CML                               & VPD                              & HTER (\%) $\downarrow$             & AUC $\uparrow$                      & HTER (\%) $\downarrow$             & AUC $\uparrow$                      & HTER (\%) $\downarrow$             & AUC $\uparrow$                      & HTER (\%) $\downarrow$             & AUC $\uparrow$                      & HTER (\%) $\downarrow$             \\ \midrule
{\color[HTML]{009901} \ding{51}} & {\color[HTML]{FE0000} \ding{55}} & {\color[HTML]{FE0000} \ding{55}} & {\color[HTML]{FE0000} \ding{55}} & 4.04                               & 99.13                               & 5.00                               & 98.89                               & 6.57                               & 98.45                               & 6.09                               & 98.12                               & 5.43                               \\
{\color[HTML]{009901} \ding{51}} & {\color[HTML]{009901} \ding{51}} & {\color[HTML]{FE0000} \ding{55}} & {\color[HTML]{FE0000} \ding{55}} & 2.86 {\color[HTML]{009901}(-1.18)} & 99.57 {\color[HTML]{009901}(+0.44)} & 3.89 {\color[HTML]{009901}(-1.11)} & 99.30 {\color[HTML]{009901}(+0.41)} & 5.00 {\color[HTML]{009901}(-1.57)} & 98.73 {\color[HTML]{009901}(+0.28)} & 2.22 {\color[HTML]{009901}(-3.87)} & 99.63 {\color[HTML]{009901}(+1.51)} & 3.49 {\color[HTML]{009901}(-1.94)} \\
{\color[HTML]{009901} \ding{51}} & {\color[HTML]{009901} \ding{51}} & {\color[HTML]{009901} \ding{51}} & {\color[HTML]{FE0000} \ding{55}} & 4.29 {\color[HTML]{009901}(-0.25)} & 98.66 {\color[HTML]{FE0000}(-0.47)} & 3.44 {\color[HTML]{009901}(-1.56)} & 99.01 {\color[HTML]{009901}(+0.12)} & 3.71 {\color[HTML]{009901}(-2.86)} & 99.17 {\color[HTML]{009901}(+0.72)} & 1.59 {\color[HTML]{009901}(-4.50)} & 99.78 {\color[HTML]{009901}(+1.66)} & 3.26 {\color[HTML]{009901}(-2.17)} \\
{\color[HTML]{009901} \ding{51}} & {\color[HTML]{009901} \ding{51}} & {\color[HTML]{009901} \ding{51}} & {\color[HTML]{009901} \ding{51}} & 2.86 {\color[HTML]{009901}(-1.18)} & 99.41 {\color[HTML]{009901}(+0.28)} & 2.67 {\color[HTML]{009901}(-2.33)} & 99.27 {\color[HTML]{009901}(+0.38)} & 1.29 {\color[HTML]{009901}(-5.28)} & 99.90 {\color[HTML]{009901}(+1.45)} & 1.49 {\color[HTML]{009901}(-4.60)} & 99.80 {\color[HTML]{009901}(+1.68)} & 2.08 {\color[HTML]{009901}(-3.35)} \\ \bottomrule
\end{tabular}%
}
\vspace{-0.3cm}
\end{table*}

\subsubsection{Single-source evaluation (P4)}

For \textbf{P4}, we strictly follow the setup in \cite{flip} to construct 12 scenarios using a single-source-to-single-target approach, leveraging the ICMO datasets. As shown in \Cref{tab:p4}, the proposed CPG-PAD surpasses suboptimal method CDFTN-L \cite{cdftn} on nine of twelve settings with an average HTER of 5.33\%, lower than half of CDFTN-L. Since it's not fair to leverage pretrained models, we trained CPG-PAD with an additional CelebA-Spoof dataset to compare with FLIP-MCL* (both leverage pretrained models). The results indicate that CPG-PAD can outperform FLIP-MCL* with an average HTER of 3.36\% demonstrating its strong generalizability on single source scenarios. 

\subsubsection{Unknown PAI evaluation (P5)}

We evaluate the generalizability of the CPG-PAD for the challenging scenario of unknown PAI, including 3D masks (i.e. silicone masks, transparent masks and mannequin head) and make-up (obfuscation, impersonation and cosmetic). For this purpose, we follow the leave-one-out protocol in \cite{foundation_model_pad} using SiW-Mv2 \cite{siwmv2} database: thirteen PAI species are used for training and the remaining PAI species is tested. \Cref{tab:pai} reports in compliance with ISO/IEC 30107-3 and benchmarks against the SiW-Mv2 baseline in terms of HTER and BPCER.

Compared with our CLIP baseline, CPG-PAD achieves clear improvements under the unknown-PAI protocol on SiW-Mv2, reducing the average HTER from 3.03\% to 1.02\% and the average BPCER100 from 11.08\% to 6.65\%. The gains are particularly evident on attack instruments with stronger appearance variation or localized attack cues, such as obfuscation make-up, cosmetic make-up, Funny Eye covering, transparent mask, and print attacks. For example, the HTER on obfuscation decreases from 14.79\% to 0.77\%, while the BPCER100 on cosmetic make-up drops substantially from 61.69\% to 22.61\%. These results indicate that concept guidance is especially beneficial for unseen PAIs involving facial region modification, occlusion boundaries, and material inconsistencies.

\subsection{Ablation Studies}

We conduct ablation studies on the same \textbf{P1} benchmark and report results across all scenarios to demonstrate the effectiveness of each component of CPG-PAD as well as the influence of key hyperparameters.

\subsubsection{Effect of CPG-PAD components}\label{ablation1}

To investigate the effect of each component in CPG-PAD, we gradually add them to the baseline CLIP and report the results of \textbf{P1} in \Cref{tab:ablation1}. The differences of each component are reported in parentheses, with green denoting improvements and red denoting degradations. The CLIP baseline in \Cref{tab:ablation1} refers to a CLIP-based PAD classification baseline with fixed class prompts, a frozen text encoder, and visual-side adaptation through a lightweight Convpass adapter, rather than a fully finetuned CLIP trained with contrastive loss. On the base of CLIP, we first add Multiple Learnable Prompts (MLP) to encourage the model to freely discover different sub-direction features that limitedly improve the performance. Adding MLP improves average HTER from 5.43\% to 3.49\% since it provides the model with greater flexibility. Then we add the Concept Mapping Loss (CML) with a simple 2D convolution layer after the visual encoder. This makes a further step to improve the performance since the auxiliary supervision of concept-associated heatmaps helps model to learn more generalizable features. However, it can not fully leverage the visual concept information since the decoder is only a simple 2D convolution layer. Finally, we change the simple decoder with the proposed Visual-Prompt Decoder (VPD) to make full use of the concept-associated heatmaps, which improve the average HTER to 2.08\%. This allows the model to effectively align the textual prompts with visual concepts, thereby enhancing the generalizability of PAD task. While adding different components of CPG-PAD to baseline CLIP, HTER and AUC is improved on almost all scenarios (only one except), indicating the effectiveness of each component proposed in CPG-PAD. 

\subsubsection{Effect of Learnable Prompt Number}

We report the result of \textbf{P1} with different $J$ in \Cref{tab:ablation2}. Since the concept number $K$ is set to 15 empirically, there are only 15 heatmaps for each sample. Thus, the number of learnable prompt $J$ can't exceed $K$. We uniformly select values in the range of 3 to 15 to analyze the impact. As $J$ increases, the model is able to learn and exploit a greater number of concepts concurrently. In \Cref{tab:ablation2}, we come up with the conclusion that the optimal value of $J$ is 15 which utilizes all visual concepts. Moreover, larger values of $J$ correspond to better performance, indicating that the more concepts discovered by VCE are utilized, the better the results. This demonstrates the effectiveness of VCE.

\begin{table}[t]
\centering
\caption{The effect of learnable prompt number $J$. We report the HTER(\%) of different $J$. Since $J$ can not exceed the discovered concept number $K=15$, we uniformly select values in the range of 3 to 15 and analyze their impact. We \textbf{BOLD} the best result in each column.}
\label{tab:ablation2}
\resizebox{\columnwidth}{!}{
\begin{tabular}{cccccc}
\toprule
$J$ & OCI $\rightarrow$ M & OMI $\rightarrow$ C & OCM $\rightarrow$ I & ICM $\rightarrow$ O & Avg. \\ \midrule
3 & 7.14 & 5.33 & 3.71 & 4.39 & 5.14 \\
6 & 7.14 & 6.00 & 3.57 & 3.94 & 5.16 \\
9 & 4.29 & 4.00 & 1.43 & 3.50 & 3.31 \\
12 & 4.29 & 3.33 & 2.86 & 2.16 & 3.16 \\
\rowcolor[HTML]{C0C0C0} 15 & \textbf{2.86} & \textbf{2.67} & \textbf{1.29} & \textbf{1.49} & \textbf{2.08} \\ \bottomrule
\end{tabular}%
}
\vspace{-0.3cm}
\end{table}

\begin{table}[t]
\centering
\caption{The effect of Visual-Prompt Decoder layer depth $Depth$. We report HTER(\%) setting $Depth$ to 0, 1, 2, 4, 8 and \textbf{BOLD} the best result in each column.}
\label{tab:ablation3}
\resizebox{\columnwidth}{!}{
\begin{tabular}{cccccc}
\toprule
$Depth$ & OCI $\rightarrow$ M & OMI $\rightarrow$ C & OCM $\rightarrow$ I & ICM $\rightarrow$ O & Avg. \\ \midrule
0 & 4.29 & 3.44 & 3.71 & 1.59 & 3.26 \\
1 & \textbf{2.61} & 5.89 & 2.93 & 1.62 & 3.26 \\
2 & 2.86 & 3.44 & 2.86 & 1.94 & 2.78 \\
\rowcolor[HTML]{C0C0C0} 4 & 2.86 & \textbf{2.67} & 1.29 & \textbf{1.49} & \textbf{2.08} \\
8 & 2.86 & 4.00 & \textbf{0.71} & 2.50 & 2.52 \\ \bottomrule
\end{tabular}%
}
\vspace{-0.3cm}
\end{table}

\begin{table}[t]
\centering
\caption{The effect of Concept Mapping Loss coefficient $\alpha$. We report HTER(\%) setting $\alpha$ to 0.1, 0.05, 0.01, 0.005, 0.001. We \textbf{BOLD} the best result in each column.}
\label{tab:ablation4}
\resizebox{\columnwidth}{!}{
\begin{tabular}{cccccc}
\toprule
$\alpha$ & OCI $\rightarrow$ M & OMI $\rightarrow$ C & OCM $\rightarrow$ I & ICM $\rightarrow$ O & Avg. \\ \midrule
0.1 & 4.29 & 6.67 & 5.71 & 3.81 & 5.12 \\
0.05 & \textbf{2.86} & 4.56 & 2.29 & 1.50 & 2.80 \\
\rowcolor[HTML]{C0C0C0} 0.01 & \textbf{2.86} & \textbf{2.67} & \textbf{1.29} & \textbf{1.49} & \textbf{2.08} \\
0.005 & 4.05 & 4.56 & 3.00 & 2.00 & 3.40 \\
0.001 & 4.29 & 4.00 & 5.00 & 2.44 & 3.93 \\ \bottomrule
\end{tabular}%
}
\vspace{-0.3cm}
\end{table}

\begin{table}[t]
\centering
\caption{The effect of different backbone. We report HTER(\%) using different ViT version including ViT-B/16, ViT-B/32 and ViT-L/14.}
\label{tab:ablation5}
\resizebox{\columnwidth}{!}{%
\begin{tabular}{cccccc}
\toprule
Backbone & OCI $\rightarrow$ M & OMI $\rightarrow$ C & OCM $\rightarrow$ I & ICM $\rightarrow$ O & Avg. \\ \midrule
\rowcolor[HTML]{C0C0C0} ViT-B/16 & 2.86 & 2.67 & \textbf{1.29} & \textbf{1.49} & 2.08 \\
ViT-B/32 & 4.52 & 3.89 & 9.29 & 2.68 & 5.10 \\
ViT-L/14 & \textbf{1.43} & \textbf{1.33} & 1.50 & 1.50 & \textbf{1.44} \\ \bottomrule
\end{tabular}%
}
\vspace{-0.3cm}
\end{table}

\begin{table}[!t]
\centering
\caption{The effect of frame number $r$ in VCE. We reproduce VCE process and report HTER(\%) by using different settings (r=1,r=2,r=3) on P1 protocol.}
\label{tab:ablation6}
\resizebox{\columnwidth}{!}{%
\begin{tabular}{cccccc}
\toprule
r & OCI $\rightarrow$ M & OMI $\rightarrow$ C & OCM $\rightarrow$ I & ICM $\rightarrow$ O & Avg. \\ \midrule
1 & \textbf{2.86} & \textbf{2.67} & 1.43 & 1.69 & 2.16 \\
\rowcolor[HTML]{C0C0C0} 2 & \textbf{2.86} & \textbf{2.67} & \textbf{1.29} & \textbf{1.49} & \textbf{2.08} \\
3 & \textbf{2.86} & 2.78 & 1.36 & 1.69 & 2.17 \\ \bottomrule
\end{tabular}%
}
\vspace{-0.3cm}
\end{table}

\subsubsection{Effect of Visual-Prompt Decoder Depth}

Visual-Prompt Decoder (VPD) plays a crucial role in the Prompt-based Concept Injection (PCI) process (shown in \Cref{tab:ablation1}). To investigate its impact on cross-domain performance, we experimented with different $Depth$ of VPD layers, where 0 layers indicate replacing VPD with a simple 2D convolution layer. We observed that increasing the number of layers improves performance when $Depth$ is below 4; however, when $Depth$ is set to 8, the performance drops. Our analysis suggests that an overly complex network may make the decoder too strong, preventing useful information in the heatmaps from being effectively back-propagated to the learnable prompts, thereby leading to degraded performance. Moreover, an 8-layer network also introduces a significant increase in parameters and computational costs. Therefore, setting $Depth$ to 4 offers the best trade-off, achieving optimal performance without excessive computational burden.

\subsubsection{Effect of Concept Mapping Loss Coefficient}

The coefficient of Concept Mapping Loss (CML) is also an important hyper-parameter of CPG-PAD method. The result of different coefficient $\alpha$ is reported in \Cref{tab:ablation4}. A small $\alpha$ (0.005 and 0.001) dilutes the effect of CML with the basic cross-entropy loss, leading to performance degradation. On the other hand, a large $\alpha$ (0.1) also results in degraded performance. This is because the concept-associated heatmaps are extracted from the visual feature space of the pretrained CLIP, whereas in the PAD task, the optimal visual space may not be perfectly aligned with that of pretrained CLIP. A large $\alpha$ forces the model to over-align with the pretrained CLIP, which ultimately harms performance. Therefore, an intermediate value of $\alpha=0.01$ proves to be the optimal choice.

\subsubsection{Effect of Backbone}

To study the effect of different backbone version, we use different ViT version as backbone including ViT-B/16, ViT-B/32 and ViT-L/14. The results are shown in \Cref{tab:ablation5}. We can draw two conclusions from the results: (1) The best performance is achieved when using ViT-L/14 as the backbone, which is consistent with expectations since ViT-L/14 ($\sim$304M parameters) has significantly more capacity than ViT-B/16 ($\sim$86M). (2) Using ViT-B/32 leads to a noticeable performance degradation. This is mainly because, given a 224×224 input, it produces only 7×7 patches, which is much coarser compared to the 14×14 patches in ViT-B/16, resulting in a loss of fine-grained information.

\subsubsection{Effect of VCE Frame Number}

In the concept discovery stage of VCE, we sample $r$ frames per video to discover visual concepts. To investigate the impact of $r$, we conduct experiments with $r=1,2,3$, and report the results in \Cref{tab:ablation6}. The results show that the best value is $r=2$. Using fewer frames may lead to information loss, while incorporating more frames can introduce redundant information and noise interference. Across different values of $r$, CPG-PAD demonstrates stable performance with only a marginal variation.

\begin{figure}[t]
\centering
\includegraphics[width=0.9\columnwidth]{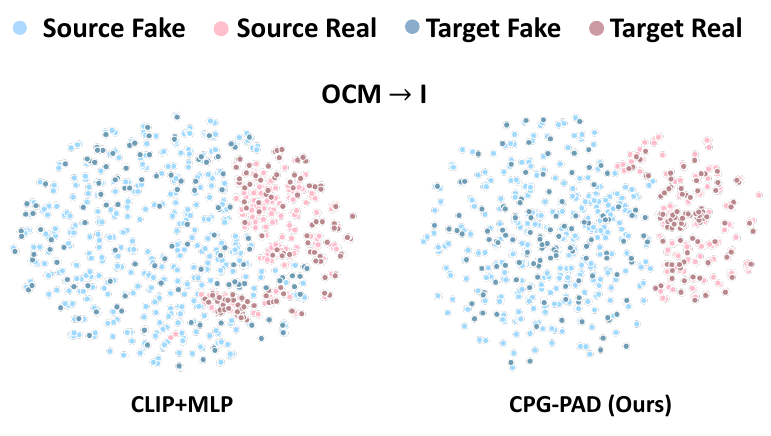}
\caption{T-SNE visualization of OCM $\rightarrow$ I in \textbf{P1} benchmark. We show T-SNE visualization of CLIP+MLP baseline and our proposed method CPG-PAD.}
\label{fig:vis}
\end{figure}

\subsection{Visualization and Analysis}

We compare CPG-PAD against the degraded model (CLIP+MLP) through T-SNE \cite{tsne} visualization on OCM $\rightarrow$ I scenarios in \textbf{P1}. As shown in \Cref{fig:vis}, the CLIP+MLP baseline produces feature distributions where red (real) and blue (fake) samples are intermixed, and the dark-colored target samples are scattered widely within each class, reflecting large intra-class variation and unclear decision boundaries. By contrast, CPG-PAD forms two visibly distinct clusters of real and fake samples, and the target samples (dark points) are tightly grouped around their source counterparts (light points), showing improved compactness and cross-domain alignment.

\begin{table}[t!]
\centering
\caption{Computational overhead. We compare CPG-PAD with CLIP.} 
\label{tab:computation_cost}
\resizebox{\columnwidth}{!}{
\begin{tabular}{c|cccc}
\toprule
Model &
\begin{tabular}[c]{@{}c@{}}Training Time\\ (batch=24)\end{tabular} &
\begin{tabular}[c]{@{}c@{}}Inference Time\\ (batch=64)\end{tabular} &
\begin{tabular}[c]{@{}c@{}}Training Memory\\ (batch=24)\end{tabular} &
\begin{tabular}[c]{@{}c@{}}Inference Memory\\ (batch=64)\end{tabular} \\ \midrule
CLIP & 0.10s & 0.07s & 3196M & 1490M \\
CPG-PAD & 0.12s & 0.08s & 6708M & 1632M \\ \bottomrule
\end{tabular}%
}
\vspace{-0.3cm}
\end{table}

\subsection{Computational Overhead}

Table~\ref{tab:computation_cost} shows the computational time for training and inference on a single RTX 3090 GPU, the results show that CPG-PAD is not a heavy burden for both training and inference. The comparison between CPG-PAD and CLIP demonstrates the applicability of CPG-PAD in real-world scenarios.

\section{Conclusion}

In this paper, we introduce Concept-informed Prompts Guided PAD (CPG-PAD), a novel framework that enhances the generalizability of PAD models by incorporating visual concepts from pretrained VLMs into PAD process. Concretely, we propose Visual Concept-driven Enhancement (VCE) which can discover PAD-relevant visual concepts and enhance domain data with fine-grained feature heatmaps corresponding to each discovered concept. With the help of these concept-associated heatmaps, Prompt-based Concept Injection (PCI) encourages the model to learn multiple concept-informed prompts via the cooperation of Visual-Prompt Decoder (VPD) and concept mapping loss, allowing CPG-PAD to precisely identify generalizable features when encountering unseen target domains. Extensive cross-domain quantitative experiments and in-depth analyses demonstrate the effectiveness of the proposed CPG-PAD method.

\bibliographystyle{IEEEtran}
\bibliography{ref}

\end{document}